\documentclass[10pt,twocolumn,letterpaper]{article}

\usepackage[pagenumbers]{cvpr}

\newcommand{\project}{ReScene4D}
\newcommand{\Circled}[1]{\raisebox{.5pt}{\textcircled{\raisebox{-.9pt} {\scriptsize #1}}}}

 \usepackage{algorithm}
 \usepackage{colortbl}
 \usepackage{svg}
\usepackage{amsmath,amssymb}
\usepackage{graphicx}
\usepackage{array}
\usepackage{booktabs}
\usepackage{xcolor}
\usepackage{tikz}
\usepackage{subcaption}
\usepackage{multirow}
\usepackage{adjustbox}
\usepackage{overpic}
\usepackage[accsupp]{axessibility}
\usepackage{xstring}

\usepackage{algorithmic}

\definecolor{good}{HTML}{E1F5D8} 
\definecolor{decrease}{HTML}{FFD9D6}    
\definecolor{baseline}{HTML}{DBE2EF}    
\definecolor{improve}{HTML}{F8FAE6}   
\definecolor{palegray}{HTML}{F3F3F3}
\definecolor{m_red}{RGB}{255,209,209}
\definecolor{my_bb}{HTML}{d9ead3}
\definecolor{my_transformer}{HTML}{d9d9d9}
\definecolor{my_query_refine}{HTML}{c7dbf7}
\definecolor{my_mask}{HTML}{f3cbcb}
\definecolor{my_adapt}{HTML}{b3a7d7}

\definecolor{purple}{RGB}{99,80,162}

\DeclareRobustCommand{\colorsquare}[1]{\tikz{\path[draw=black,fill=#1, thick, rounded corners=0.6pt] (0,0) rectangle (6pt,6pt);}}

\definecolor{cvprblue}{rgb}{0.21,0.49,0.74}
\usepackage[pagebackref,breaklinks,colorlinks,allcolors=cvprblue]{hyperref}

\def\confName{CVPR}
\def\confYear{2026}

\title{ReScene4D: Temporally Consistent Semantic Instance Segmentation of Evolving Indoor 3D Scenes}

\author{
Emily Steiner\textsuperscript{1} \quad
Jianhao Zheng\textsuperscript{1} \quad
Henry Howard-Jenkins\textsuperscript{2} \quad
Chris Xie\textsuperscript{2} \quad
Iro Armeni\textsuperscript{1} \\
\textsuperscript{1}Stanford University \quad
\textsuperscript{2} Meta Reality Labs Research \\
\footnotesize\url{https://www.easteine.com/rescene4d/}
}

\begin{document}
\maketitle
\begin{abstract}
Indoor environments evolve as objects move, appear, or leave the scene. Capturing these dynamics requires maintaining temporally consistent instance identities across intermittently captured 3D scans, even when changes are unobserved. We introduce and formalize the task of temporally sparse 4D indoor semantic instance segmentation (SIS), which jointly segments, identifies, and temporally associates object instances. This setting poses a challenge for existing 3DSIS methods, which require a discrete matching step due to their lack of temporal reasoning, and for 4D LiDAR approaches, which perform poorly due to their reliance on high-frequency temporal measurements that are uncommon in the longer-horizon evolution of indoor environments. We propose \project{}, a novel method that adapts 3DSIS architectures for 4DSIS without needing dense observations. Our method enables temporal information sharing—using spatiotemporal contrastive loss, masking, and serialization—to adaptively leverage geometric and semantic priors across observations. This shared context enables consistent instance tracking and improves standard 3DSIS performance. To evaluate this task, we define a new metric, t-mAP, that extends mAP to reward temporal identity consistency. \project{} achieves state-of-the-art performance on the 3RScan dataset, establishing a new benchmark for understanding evolving indoor scenes.
\end{abstract}

\vspace{-10pt}    
\section{Introduction}

\label{sec:intro}
Our built environments are not static—they are living scenes that evolve over time as objects are moved, replaced, or repurposed \cite{zhu_living_2024}. Capturing and understanding changes is crucial for a range of applications, including long-term spatial monitoring, facility management, and adaptive digital twins. Yet, most existing methods for 3D scene understanding assume that environments remain static or are observed continuously over time, as in autonomous driving settings. In contrast, real-world indoor environments are typically observed sparsely over a long horizon—across days, months, or even years—undergoing substantial semantic and geometric change between captures.

Recent advances in 3D semantic instance segmentation (3DSIS) have made it possible to identify and segment objects in static scenes with impressive accuracy \cite{schult_mask3d_2023,sun_superpoint_2023,lu2025relation3d}. However, these methods process each observation independently, neglecting temporal identity continuity. Conversely, 4D semantic segmentation and tracking methods for outdoor LiDAR data \cite{marcuzzi_mask4d_2023,yilmaz_mask4former_2024} focus on densely sampled sequences with minimal scene change, where temporal correspondences can be assumed and tracked explicitly with optical flow. When observations are temporally sparse, these assumptions no longer hold. The challenge, then, is to maintain consistent instance identities over time despite substantial changes in appearance, pose, or even topology.

\begin{figure}[t]
  \centering
  \label{fig:teaser}
  \includegraphics[width=1\linewidth]{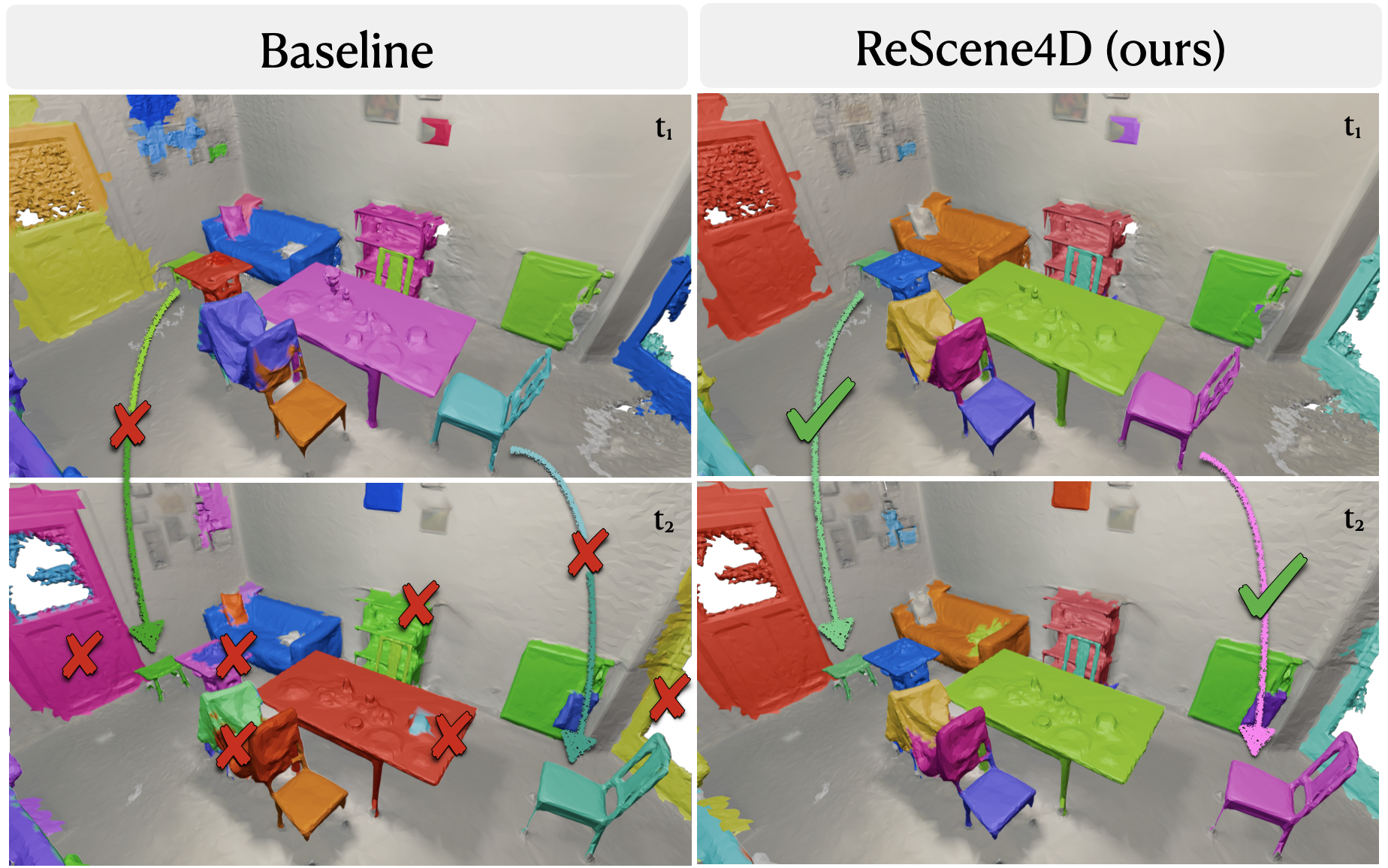}
   \caption{\textbf{\project{} on 4D Semantic Instance Segmentation}. Our method outperforms baselines which are unable to accurately assign instance identities of both static and changing objects, across multiple temporal observations of a scene (left). Our method (right) maintains instance identities between the observations, even when instances move or change. }
   \label{fig:onecol}
   \vspace{-15pt}
\end{figure}

Existing approaches only partially address this challenge. Change detection highlights where scenes differ but does not establish semantic or instance-level correspondences \cite{langer_robust_2020,adam_has_2023}. Object relocalization and reconstruction methods \cite{zhu_living_2024} can re-identify specific objects but are not designed for comprehensive scene-level reasoning. Furthermore, no existing metric jointly evaluates segmentation quality and identity consistency when observations are intermittent. Metrics such as mAP~\cite{fleet_simultaneous_2014} and LSTQ~\cite{aygun_4d_2021} measure spatial segmentation and tracking, but neither enforces strict identity consistency across all observations nor reflect identity ambiguity of sets of identical objects in temporally sparse sequences.

In this work, we introduce the task of \textit{temporally sparse 4D indoor semantic instance segmentation} (4DSIS)—the joint problem of segmenting, identifying, and temporally associating object instances across intermittently observed 3D reconstructions of evolving indoor scenes. This problem is distinct from dense LiDAR-based 4DSIS settings, characterized by intermittently acquired 3D observations over long time horizons, as indoor environments are continually constructed, modified, and used—making continuous sensing impractical and unnecessary in real-world deployments. 
We propose a new metric that extends mAP to the temporal domain, rewarding consistent instance tracking and penalizing identity switches, merges, and fragmentation. Uniquely, our metric explicitly accommodates ambiguous object identities, such as symmetric object swaps.

To tackle this problem, we propose \textbf{\textit{\project{}}}. This novel method achieves temporally consistent instance segmentation across sparsely captured 3D scans of the same indoor scene as it evolves over time. Our approach adapts existing 3DSIS architectures to the 4D setting without strictly enforcing geometric overlay or frame-to-frame alignment assumptions, unlike many LiDAR-based pipelines. We explore multiple strategies for sharing temporal information across observations and evaluate their interaction with three distinct feature backbones: a light-weight voxel-based encoder \cite{choy_4d_2019}, a 3D self-supervised geometric encoder pretrained on large-scale data \cite{wu_sonata_2025}, and a cross-modal 2D-3D self-supervised encoder pretrained on large-scale data \cite{zhang_concerto_2025}. Experiments on the 3RScan dataset \cite{wald_rio_2019} demonstrate the effectiveness of our approach and highlight both the promise and current limitations of modeling living indoor environments. Our contributions are threefold:
\begin{itemize}
    \item We define and formalize the task of temporally sparse 4D semantic instance segmentation (4DSIS) and introduce a metric that jointly evaluates segmentation quality and temporal identity consistency.

    \item We demonstrate that sharing information across temporal observations—despite scene changes—improves not only 4DSIS but also per-stage 3DSIS. To achieve this, we propose and evaluate three temporal fusion strategies.

    \item Our method, ReScene4D, predicts temporally consistent instance queries without assuming geometrically aligned points share instance identity, disparate matching steps, predefined correspondences, or high-frequency temporal sampling. We achieve state-of-the-art performance on 3RScan, establishing a new benchmark for indoor 4DSIS.
\end{itemize}

\section{Related Work}
\label{sec:related}

\vspace{4pt}
\noindent\textbf{3D Semantic Instance Segmentation.}
Early 3DSIS methods fall into two categories: proposal-based \cite{hou_3d-sis_2019, engelmann_3d-mpa_2020, yang_learning_2019}, which detect 3D bounding boxes and predict masks within them, and grouping-based \cite{jiang_pointgroup_2020,vu_softgroup_2022,liang_instance_2021}, which aggregate per-point predictions into instances. Convolutional approaches \cite{he_dyco3d_2021,he_pointinst3d_2022,ngo_isbnet_2023} generate masks directly from features. While effective for static scenes, these methods rely on geometric relationships and intermediate predictions, making them sensitive to error propagation. Recent work like SphericalMask \cite{shin_spherical_2024} improves accuracy by leveraging geometric compactness and refinement techniques. However, these strategies struggle in 4D settings, where instances move or change over time, breaking the assumptions underlying point grouping and proposal-based approaches.

In recent years, methods have shifted to end-to-end query-based transformer architectures, inspired by MaskFormer \cite{cheng_per-pixel_2021} and Mask2Former \cite{cheng_masked-attention_2022} in 2D segmentation. Adaptations like Mask3D \cite{schult_mask3d_2023} and SPFormer \cite{sun_superpoint_2023} use a 3D Sparse-UNet backbone with a masked-attention transformer to predict instance masks and semantic labels from learned queries. Subsequent works—such as QueryFormer \cite{lu_query_2023}, Competitor \cite{wang_competitorformer_2025}, Maft \cite{lai_mask-attention-free_2023}, and Relation3D \cite{lu2025relation3d}—improve query distribution, convergence, spatial reasoning, and feature refinement. These methods collectively demonstrate the advantages of transformer-based queries over traditional proposal- or grouping-based approaches, while recent large-scale self-supervised point cloud encoders such as Sonata \cite{wu_sonata_2025} and Concerto \cite{zhang_concerto_2025} have achieved state-of-the-art results across a range of 3D scene understanding tasks.

Our aim is not to optimize queries or backbone designs, but to develop  strategies and mechanisms for extending well-established spatial architectures to the temporal dimension—emphasizing the importance of jointly optimizing temporally consistent predictions and developing architectural elements that support temporal information sharing. As our experiments show, stronger pretrained 3D backbones improve performance on the 4DSIS task; in the absence of large-scale 4D indoor datasets, relying solely on purpose-built 4D architectures is insufficient to solve the problem.

\vspace{4pt}
\noindent\textbf{4D LiDAR Panoptic Segmentation.}
For LiDAR Panoptic Segmentation, the data is characterized by long sequences of densely observed spatiotemporal point cloud measurements of outdoor scenes. Early 4D LiDAR segmentation methods ~\cite{ aygun_4d_2021, hong_lidar-based_2022, kreuzberg_4d-stop_2022, zhu_4d_2023, marcuzzi_contrastive_2022} cluster points by modeling object centers across superimposed scans, assuming minimal per-frame motion or grouping learned features. These methods rely on spatiotemporal clustering, geometric proximity, and handcrafted motion models.

Recent 4D LiDAR panoptic segmentation methods~\cite{stratil_new_2025} adopt transformer-based, query-driven frameworks to process dynamic outdoor point cloud sequences with unified architectures. Mask4Former~\cite{yilmaz_mask4former_2024} and SP2Mask4D~\cite{park_sp2mask4d_2025} operate on superimposed scans to enforce spatial alignment, an assumption suitable for LiDAR with minute frame-wise changes but restrictive when more significant scene changes occur between observations. Mask4D~\cite{marcuzzi_mask4d_2023} instead propagates instance queries across consecutive scans but lacks explicit temporal information sharing, preventing later scans from correcting earlier predictions or aligning instance features in a shared context. 
Our approach introduces a unified end-to-end architecture for 4DSIS that jointly refines spatio-temporal instances by sharing semantic information across time and without relying on dense observations.

\vspace{4pt}
\noindent\textbf{Re-Seen Changing Scenes.}
Understanding long-term changing 4D scenes, which are sparsely sampled over time, contrasts with 4D high-frequency short-term dynamic scenes. To capture long-term scene changes, datasets such as 3RScan~\cite{wald_rio_2019}, ReScan~\cite{halber_rescan_2019}, MultiScan \cite{mao_multiscan_2022} and NSS~\cite{sun_nothing_2025} have been introduced. Here, object appearance and arrangement between scans can change drastically, rendering motion models, flow, and tracking-based pipelines ineffective.

Several tasks have emerged around understanding from these datasets. Prior works on change detection \cite{langer_robust_2020, avidan_objects_2022}, changed object pose estimation \cite{adam_has_2023}, or object relocalization \cite{ wald_rio_2019} address these changes through sequential pipelines—detecting, matching, and labeling object differences—but rely on externally predicted masks and assume consistent instance boundaries. 
MORE~\cite{zhu_living_2024} jointly addresses reconstruction and re-localization to leverage repeated scans for improved geometry and appearance, but still relies on ground-truth mask filtering and heuristic matching to assign instance identities.

Despite increasing attention to evolving 3D environments and the common reliance on disparate association or matching steps, no unified formulation currently exists for semantic instance segmentation that \emph{jointly integrates and refines} information from multiple indoor temporal observations. For example, RescanNet~\cite{halber_rescan_2019} operates inductively, incrementally associating object instances across sparse rescans, initializing from ground-truth segmentations and using handcrafted segmentation, registration, and matching pipelines. In contrast, our method jointly predicts temporally consistent instance masks without requiring prior segmentation, motion assumptions, or correspondence heuristics. This unified formulation provides the representational foundation for downstream change detection, relocalization, and reasoning tasks, enabling a coherent understanding of how real-world environments evolve.

\section{\project{}}

\paragraph{Problem Definition}
We formally define \textit{Temporally Sparse 4D Semantic Instance Segmentation}  (4DSIS): Given a \textit{sequence} of temporally distinct 3D scans $\mathcal{P} = \{P^{(1)}, ..., P^{(T)}\}$ of a scene $\mathcal{S}$, where each scan $P^{(t)}$ represents a single temporal \textit{stage} $t$ and contains $N_t$ points, in which some unobserved changes have occurred, the objective is to jointly predict  all points across all scans into $K$ instance masks $\mathcal{M} = \{m^1, ..., m^K\}$. Each mask $m^k \in \{0, 1\}^{N}$ spans the entire sequence, where $N = \sum_{t=1}^{T} N_t$, covering points from any combination of stages and is assigned a single semantic classification label $c$. 

\subsection{Preliminaries}
\begin{figure*}[t]
  \centering
  \includegraphics[width=1\linewidth]{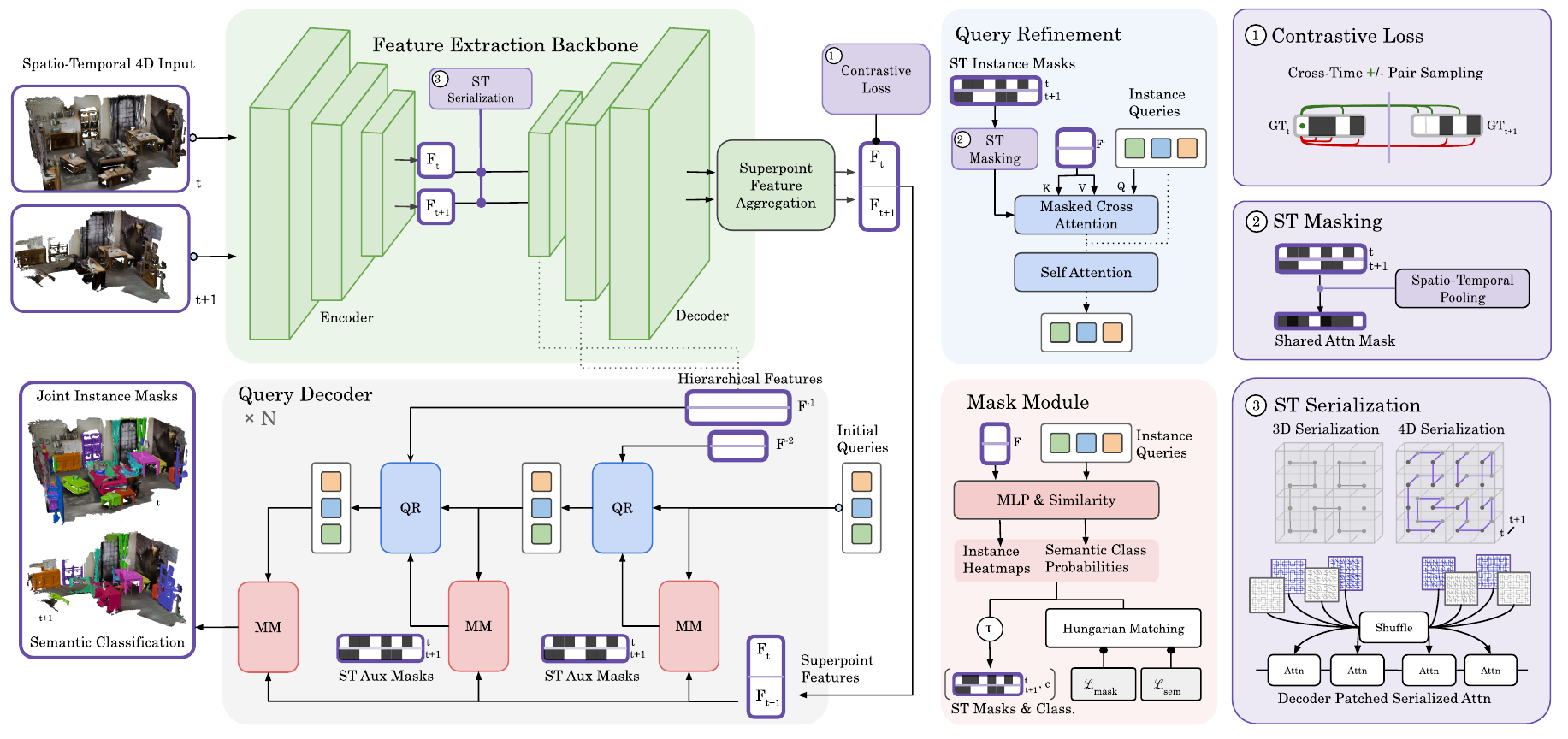}
   \caption{\textbf{Overview of \project{} Architecture.} Given a temporal sequence of $T$ 3D observations, hierarchical features for each temporal stage, preserving temporal distinction are extracted using a backbone encoder~\colorsquare{my_bb}. A transformer-based query decoder~\colorsquare{my_transformer} iteratively refines spatio-temporal (ST) instance queries by jointly sampling across temporal hierarchical features~\colorsquare{my_query_refine}. Given ST superpoint features and ST queries, the mask module~\colorsquare{my_mask} predicts joint binary masks and semantic classes consistent across the sequence. Adaptations for 4DSIS are denoted in \textcolor{purple}{\textbf{purple}}. Our temporal information sharing modules \Circled{1}, \Circled{2}, \Circled{3 } facilitate cross-temporal consistency and shared learning via cross-time contrastive loss, ST mask pooling, and ST decoder serialization~\colorsquare{my_adapt}.}
   \label{fig:overview} \vspace{-10pt}
\end{figure*}

We adapt the mask‑transformer architecture from Mask3D \cite{schult_mask3d_2023} to 4DSIS. For clarity, an overview of key components is summarized here and visualized alongside our method’s specifics in Fig.~\ref{fig:overview}.  Mask3D comprises a feature backbone (\colorsquare{my_bb}) paired with a query decoder (\colorsquare{my_transformer}) that includes query refinement(\colorsquare{my_query_refine}) and mask modules (\colorsquare{my_mask}). The architecture maintains \textit{instance queries}, each representing a potential object instance in the point cloud. These queries are iteratively refined at multiple hierarchical feature resolutions: through masked-cross-attention to backbone features, only attending to foreground regions as indicated by auxiliary mask predictions from previous steps, and self-attention among queries. This process, repeated over several transformer layers, progressively aligns and strengthens query representations while mitigating competition. In the final stage, the mask module (\colorsquare{my_mask}) uses the refined queries to predict binary instance masks and semantic classes for each detected object, leveraging similarity to semantic point cloud features (more details in Supp.).

\subsection{Spatio-Temporal Architecture}
An overview of our 4DSIS approach is shown in Fig.~\ref{fig:overview}.

\vspace{4pt}
\noindent\textbf{Spatio-Temporal 4D Input.}
We represent a sequence of point cloud scans from the same scene as a unified registered spatio-temporal 4D point cloud, \(\mathcal{P} \in \mathbb{R}^{N \times 4}\) of size N. While many LiDAR-based methods~\cite{stratil_new_2025, park_sp2mask4d_2025, yilmaz_mask4former_2024}  utilize pose estimates to align scans within a global coordinate frame condensed to 3 dimensions (removing the temporal dimension distinction), our approach keeps points from different temporal observations separate and does not pool the temporal coordinate during voxelization into $\mathcal{V} \in \mathbb{Z}^{K_0 \times 4}$. 

\vspace{4pt}
\noindent\textbf{Feature Extraction Backbone.} 
We evaluate our method with two distinct backbone architectures for sparse point cloud feature extraction. Both process the voxelized spatio-temporal point clouds \(\mathcal{V} \in \mathbb{Z}^{K_0 \times 4}\) to produce hierarchical features \(F^{(r)} \in \mathbb{R}^{M_r \times D_r}\) at different resolution layers \(r\) capturing global to fine-grained semantic information integral to the mask transformer query refinement strategy. Hierarchical features are indexed by resolution, with the full resolution level denoted \(F\) and coarser levels \(F^{-1}, F^{-2}, \ldots\). In both backbones, the encoder operates independently at each temporal stage, producing distinct feature sets \(F_t, F_{t+1}, \ldots\) for each point cloud. Superscripts index resolution and subscripts index temporal stage.

The first backbone is a commonly used sparse convolutional U-Net~\cite{choy_4d_2019} trained from scratch, with minor dimensional adjustments from the Mask3D implementation to accommodate 4D voxels.
The second is a state-of-the-art point transformer V3 (PTv3)~\cite{wu_point_2024}, a self-supervised encoder pre-trained on large-scale data, used in two variations: one trained on 3D data only (Sonata \cite{wu_sonata_2025}) and another on joint 2D-3D data (Concerto \cite{zhang_concerto_2025}), both of which yield strong semantic point cloud features validated on diverse 3D benchmarks. We aim to leverage these recent advances for semantic robustness and temporal consistency. PTv3 employs serialized attention, which patches the voxels with multiple serialization-based neighbor patterns to expand receptive fields efficiently. In all our experiments, we freeze the encoder and treat each temporal stage independently within a batch, restricting serialization patterns to individual 3D scenes. The backbone decoder is trained from scratch but uses different serialized attention blocks (see Sec.~\ref{sec:sharing_temp_info}). 

\vspace{4pt}
\noindent\textbf{Masked Transformer Architecture Adaptations.}
Following Mask4Former~\cite{yilmaz_mask4former_2024}, we use spatio-temporal queries to jointly predict masks for all stages of a sequence, removing the need for separate matching steps. Queries are shared across temporal stages, producing a set of masks per stage while maintaining consistent instance correspondence. To address domain shift between panoptic 4D LiDAR segmentation and our spatio-temporal input, as with our 4D point cloud input, we treat superpoints as independent at each time step, without temporal accumulation. As a result, query positional encodings use 4D Fourier features for spatiotemporal coordinates $(x, y, z, t)$, normalized per hierarchy level and projected by a shared Gaussian matrix~\cite{tancik_fourier_2020}.

A challenge of joint-4DSIS is the computational cost. Our approach enables efficient 4D reasoning through several design choices. While joint query refinement attends to backbone features across all temporal stages, the number of queries remains fixed regardless of sequence length, promoting object matching while keeping memory usage low. Hierarchical features are downsampled and padded for the query-refinement cross-attention block, with sampling independent of sequence length to further reduce the additional training cost associated with multi-stage sequences. 

To ensure temporal consistency, unmatched predictions incur a higher no-object semantic loss (\(\lambda_{no obj} = 0.2\)) than single-stage settings, discouraging duplicates across time.

\subsection{Temporal Information Sharing} 
\label{sec:sharing_temp_info}
In addition to our 4D architectural changes, we evaluate three methods for sharing information across temporal observations. While we do not enforce spatially coincident voxels to share labels, spatial cues and repeated instances still provide useful signals. Our approach shares temporal information by \textit{flexibly} leveraging both geometric and semantic priors. This design targets key 4DSIS challenges, including partial scans, noise, unobserved changes, and small registration misalignment, relaxing the common assumption and reliance on incremental frame-wise scene changes while effectively exploiting spatio-temporal context.

\vspace{4pt}
\noindent\textbf{1. Contrastive Loss.}
To further enhance instance discrimination and propagate temporal information across features, we introduce a supervised contrastive learning mechanism on pooled superpoint features. This enforces semantic consistency of superpoint-level representations within the same instance, promotes separation between different instances, and simultaneously propagates temporal cues via supervision signals from both positive- and negative-pair sampling across all time frames. This encourages the network to learn temporally-consistent feature representations.

Given $S$ superpoint features pooled from point-level embeddings, we construct a binary relation matrix $R_{GT} \in \{0,1\}^{S \times S}$ using instance annotation across the entire temporal sequence:
\begin{equation}
    R_{GT}(i, j) =
    \begin{cases}
        1, & \text{if } i, j \text{ belong to the same instance,} \\
        0, & \text{otherwise,}
    \end{cases}
    \label{eq:seg_relation}
\end{equation}
where $i$ and $j$ index different pooled superpoints. Self-similarities and superpoints without positives are excluded.

We implement the contrastive loss using the InfoNCE formulation~\cite{oord_representation_2019, chen_simple_2020}.
For multiple positive targets, we compute the supervised InfoNCE contrastive loss as:
\begin{equation}
    \mathcal{L}_{\text{cont}} = - \frac{1}{|S^+|} \sum_{i \in S^+} \log \frac{\sum_{j \in P(i)}\exp(L_{ij})}{\sum_{k} \exp(L_{ik})}
    \label{eq:seg_infonce}
\end{equation}
where $P(i)$ denotes the set of positive indices for anchor $i$ (as defined by $R_{GT}$), $S^+$ is the set of anchors with at least one positive and $L_{ij}$ is the normalized cosine similarity. More details in Supp.

\vspace{4pt}
\noindent\textbf{2. Spatio-Temporal Masking.}
To enable different temporal stages to guide each other, we incorporate spatio-temporal masking into the masks used for query refinement via masked cross-attention. Specifically, we temporally pool the masks—by performing a logical OR operation to capture sparse voxel overlaps—at the corresponding hierarchical level as visualized in Fig.~\ref{fig:overview}. As a result, different stages can guide queries to attend to the same spatio-temporally aligned locations. At coarser resolution levels, voxels are more likely to overlap, which promotes temporal information sharing; as the resolution becomes finer, overlaps between voxels decrease, naturally allowing the masks to be independently refined at each stage, despite their initial sharing. We do not enforce that points aligned across time must be included in the final mask prediction (or the loss), as would be the case if superimposed point clouds were strictly matched and penalized together. When there is no voxel overlap, ST-masking has no effect. 

\vspace{4pt}
\noindent\textbf{3. Spatio-Temporal Decoder Serialization.}
For the PTv3 based backbones Concerto and Sonata \cite{wu_point_2024, zhang_concerto_2025, wu_sonata_2025}, the encoder and decoder use point cloud serialization via space-filling curves (e.g., Z-order, Hilbert, and Trans variants) to order  points in 3D, enabling structured attention and efficient scaling. To support spatio-temporal information sharing we adapt the serialization strategy in the decoder, applying all four curve patterns to temporally merged point clouds. Samples from different time stages are combined without sparsifying or removing duplicate voxels,  with batch offsets adjusted so that each temporal stage can be can be processed individually or jointly as needed.

At each hierarchical decoder layer we randomly shuffle both the original spatial serialization and the spatio-temporal serialization patterns derived from the entire sequence. This design enables the decoder to attend to both spatial and temporal neighbors, enriching the neighborhood and allowing it to leverage complementary information across time steps while maintaining spatial fidelity. This method is visualized in Fig.~\ref{fig:overview}. Importantly, we keep the serialization patterns fixed for the frozen pretrained encoder, ensuring the input distribution matches the original backbone pretraining and avoiding domain shift.

\section{Evaluation Metrics}
Existing 3D semantic segmentation metrics are inadequate for this task. We propose temporal mean average precision (t-mAP), which evaluates both per-stage segmentation quality and temporal consistency.

\begin{figure}[t]
  \centering
  \begin{minipage}[t]{0.7\linewidth}
    \vspace{0pt} 
    \centering
    \includegraphics[width=\linewidth]{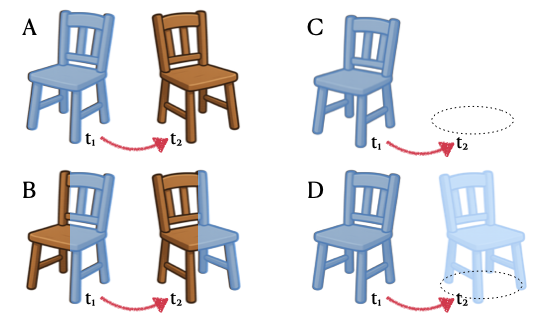}
  \end{minipage}\hfill
  \begin{minipage}[t]{0.25\linewidth}
    \vspace{4mm}
    \centering
    \scriptsize
    \setlength{\tabcolsep}{2.5pt} 
    \renewcommand{\arraystretch}{1.05}
    \begin{tabular}{@{}lcc@{}}
      \toprule
      \textbf{Case} & \textbf{IoU} & \textbf{t-IoU} \\
      \midrule
      A & 0.5 & 0.0 \\
      B & 0.5 & $\approx0.5$ \\
      C & 1.0 & 1.0 \\
      D & 0.5 & 0.0 \\
      \bottomrule
    \end{tabular}
  \end{minipage}
  \caption{\textbf{Toy Examples for Temporal Metrics.} Right: summary table showing IoU and t-IoU scores for four cases.}
  \vspace{-10pt}
  \label{fig:toy}
\end{figure}

\vspace{4pt}
\noindent\textbf{Existing metrics}
The standard metric for indoor 3D SIS, mean Average Precision (mAP), evaluates detection and segmentation using region-level IoU thresholds~\cite{fleet_simultaneous_2014}, averaged per semantic class. In 3D settings, a predicted instance is correct if its point-level IoU with a ground-truth instance exceeds the specified threshold. However, evaluating each stage independently fails to capture spatiotemporal identity continuity. Methods such as \cite{halber_rescan_2019}, propose a distinct instance transfer metric which aggregates performance across time using overall IoU. As shown in Fig.~\ref{fig:toy} (examples A and B), IoU alone cannot distinguish between high per-stage overlapping yet inconsistent temporal identities and lower-overlap predictions that preserve consistent identities. Moreover, the metric in~\cite{halber_rescan_2019} does not handle \textit{ambiguous instance groups}—sets of visually identical objects, such as furniture, that are common indoors. These groups under unobserved changes have multiple identity assignments equally valid in both annotation and prediction.

Similarly, LiDAR metrics such as LSTQ ~\cite{aygun_4d_2021} do not accommodate ambiguous instance groups and, like mAP, rely on standard IoU for the association score, which can obscure identity switches in our short, intermittent sequences (see Fig.~\ref{fig:toy}).
Furthermore, LSTQ is tailored for outdoor panoptic segmentation and was specifically designed to avoid over-emphasizing small segments and stuff classes~\cite{aygun_4d_2021}—unlike indoor 4DSIS, where tracking precise small instances is critical for applications such as robotics.

We aim to evaluate 4DSIS over sequences of length $\mathcal{T}$ using a single metric with the following properties:
\begin{enumerate}
    \item \textbf{Consistency with mAP:} Reduces to standard mAP when $\mathcal{T}=1$, while maintaining strict semantic enforcement and one-to-one matching standard for indoor SIS.
    \item \textbf{Temporal consistency:} Rewards identity continuity while penalizing switches, merges, and fragmentation.
    \item \textbf{Ambiguity tolerance:} Penalizes merges but not symmetric swaps between visually identical instances.
    \item \textbf{Sequence-length sensitivity:} Naturally decreases for longer sequences as errors accumulate, reflecting the greater challenge of maintaining temporal consistency.
\end{enumerate}

\vspace{4pt}
\noindent\textbf{Temporal mean Average Precision}
Our proposed metric, temporal mean average precision (t-mAP), enforces the IoU threshold individually at each temporal stage \(t\). We define a prediction or ground truth instance for the semantic class \(c\) as the union of the set of points comprising that instance at all temporal stages. 
\[
pr_i(c) = \bigcup_{t=1}^{T} p_i(c, t)
\quad\text{and}\quad 
gt_i(c) = \bigcup_{t=1}^{T} g_i(c, t)
\]

For a ground truth instance \(gt_i(c)\), a predicted instance \(pr_i(c)\) is a positive detection for a set IoU threshold \(\tau\) if:
\[ \text{t-IoU}(p_i(c), g_i(c)) :=  \min_{t \in [1,T]} \{\text{IoU}(p_i(c, t), g_i(c, t))\} > \tau\]

For multi-stage instances, stages where both \(g_i(c, t)\) and \(p_i(c, t)\) are absent are excluded from t-IoU. If \(p_i(c, t)\) exists but \(g_i(c, t)\) does not, then \(\mathrm{IoU}(p_i(c, t), g_i(c, t)) = 0\). See Fig.\ref{fig:toy} (examples C, D) for edge cases.

A distinctive feature of t-mAP is its handling of ambiguous ground-truth instances. While prediction inconsistency within an ambiguous group is not penalized, each prediction must be uniquely assigned. To enforce this, we use an iterative assignment strategy to pseudo-disambiguate ground-truth instances (defined in the Supp.). For an ambiguous group \(\mathcal{G}\) with \(n_{amb}\) instances across \(T\) stages, predictions overlapping any instance form a set \(\mathcal{P}\). We construct a weight matrix \(W\), where each entry combines the IoU and the prediction confidence \(p _ {k} ^ {t} \) for instance \(k\) at stage \(t\). Assignments proceed by selecting the highest-weighted prediction at each step; stages without a matching prediction remain unassigned until all votes are counted. This process ensures unique and consistent mapping even when some components lack sufficient coverage or are already claimed.

True and false detections use temporally enforced IoU and matching, following standard mAP~\cite{fleet_simultaneous_2014}. We focus on t-mAP, but the same definitions apply to t-mPrec and t-mRec. As t-mAP decreases with more temporal steps \(T\), \(T\) should be reported, defaulting to \(T=2\).

\section{Experiments}
\label{sec:results}

\noindent\textbf{Datasets and Metrics.}
We evaluate on the \textbf{3RScan} dataset \cite{wald_rio_2019}, that contains 478 unique real-world indoor environments captured multiple times, totaling 1,428 registered RGB-D scans with temporally consistent instance-level semantic annotations. For each environment, we construct length-2 temporal sequences by randomly ordering the primary scan and its rescans and applying a sliding window.

To improve semantic coverage, we additionally incorporate \textbf{ScanNet}~\cite{dai_scannet_2017,rozenberszki_language-grounded_2022} into training. 
It contains over 1,500 annotated \textit{single}-scan indoor scenes spanning diverse environments and object categories. Since our model does not require explicit temporal bottlenecks, it can process both single- and multi-stage inputs. We therefore build a mixed training set by sampling two-stage sequences from 3RScan and single-scan scenes from ScanNet with a 1.0:0.8 ratio. Evaluations follow the shared NYU40 semantic label set.

Evaluation is conducted using our proposed t-mAP, a sequence-level mean Average Precision (mAP) metric, as well as temporally enforced mean Recall (t-mRec) for different object change types.

\vspace{4pt}
\noindent\textbf{Baselines.}
Since there are no established indoor 4DSIS baselines, we compare our approach to Mask4D~\cite{marcuzzi_mask4d_2023}, Mask4Former~\cite{yilmaz_mask4former_2024}, and the 3D semantic instance segmentation baseline Mask3D~\cite{schult_mask3d_2023} augmented with sequence matching.
\textbf{Mask4D} and \textbf{Mask4Former} were designed for outdoor LiDAR-based 4D panoptic segmentation and operate exclusively on sequence input, i.e., expect T $>$ 1; training them on single-stage scenes would require non-trivial modifications to the training pipelines. We therefore maintained their configurations to ensure a faithful and reproducible baseline comparison. Consequently, we train them only on 3RScan sequences. We initialize Mask4Former with a ScanNet-pretrained Minkowski backbone from Mask3D. In contrast, Mask4D’s MaskPLS backbone is trained from scratch on 3RScan, resulting in notably degraded performance due to its LiDAR-specific design.  We report results of our method with these training settings in the Supp. 

We train Mask3D on single scans sampled from both 3RScan and ScanNet, with dataset weighting equivalent to our temporal sequence experiments. For sequence matching, we implement two strategies:
\begin{itemize}
    \item \textit{Semantic matching}: Instance predictions are grouped by semantic prediction and aligned across scans using average instance feature similarity and Hungarian matching.
    \item  \textit{Geometric matching}: For the second temporal observation, each point is assigned to the instance of its nearest neighbor in the first scan (K-tree). 
\end{itemize}

\vspace{4pt}
\noindent\textbf{Implementation Details.} These are provided in the Supp.

\subsection{4D Instance Semantic Segmentation}
Evaluation on all temporal sequences for the 4DSIS task is shown in Table~\ref{tab:main_results}. For all tables, best results in \textbf{bold} and second best \underline{underlined}. For ours, we report results using three feature backbones, (Minkowski (M), Sonata (S), and Concerto (C)), each paired with different temporal information-sharing strategies, showing the best-performing configuration for each. Qualitative results are in Fig.~\ref{fig:onecol} and the Supp.

\begin{table}[t]
    \centering
    \footnotesize 
    \setlength{\tabcolsep}{3pt} 
    \begin{tabular}{l|ccc|ccc}
      \toprule
        \textbf{Method} & \textbf{tmAP} & \textbf{tmAP$_{50}$} & \textbf{tmAP$_{25}$} & \textbf{mAP} & \textbf{mAP$_{50}$} & \textbf{mAP$_{25}$} \\
      \midrule
      Mask4D~\cite{marcuzzi_mask4d_2023}         & 1.3   & 2.9  & 8.7  & 2.1   & 5.5   & 21.2 \\
      Mask4Former~\cite{yilmaz_mask4former_2024} & 17.0  & 38.9 & 59.1 & 21.7  & 45.6  & 66.3 \\
      Mask3D~\cite{schult_mask3d_2023}+sem       & 20.1  & 32.9 & 38.6 & 25.9  & 42.3  & 73.9 \\
      Mask3D~\cite{schult_mask3d_2023}+geo       & 20.7  & 43.1 & 62.4 & 29.7  & 54.1  & 70.9 \\
      \project{} (M)    & 31.6 & 49.5 & 61.6 & 39.2 & 60.7 & 74.1 \\
      \project{} (S)    & \underline{33.2}  & \underline{50.7} & \underline{63.3} & \underline{40.9}  & \underline{62.8}  & \underline{79.1} \\
      \project{} (C)    & \textbf{34.8} & \textbf{52.5} & \textbf{66.8} & \textbf{43.3} & \textbf{64.3} & \textbf{81.9} \\
      \bottomrule
    \end{tabular}
    \caption{\textbf{4DSIS Scores on the 3RScan dataset \cite{wald_rio_2019}}. We report both the t-mAP and standard mAP scores with different IoU thresholds averaged over 18 classes.}
    \label{tab:main_results}
    \vspace{-12pt}
\end{table}

All \project{} variants outperform baselines across metrics, highlighting the importance of explicit temporal information sharing for consistent instance segmentation. Notably, the temporal LiDAR baselines (Mask4D, Mask4Former) underperform compared to Mask3D with post-hoc matching, due to their reliance on assumptions like dense observations and smooth motion, which do not apply in sparse, evolving indoor scenes. Mask4D is further constrained by its LiDAR-specific backbone, which must be trained from scratch on limited 3RScan data, preventing effective use of pretrained spatial features.

Our analysis of \project{} variants highlights the impact of backbone choice and temporal strategy. The top-performing model, \project{} (C), pairs the Concerto backbone with cross-time contrastive loss and spatio-temporal serialization, demonstrating the value of semantically expressive, self-supervised encoders for faster convergence and improved temporal modeling. \project{} (S) achieves its best results with spatio-temporal serialization and masking. Although PTv3 encoders are not fully fine-tuned due to computational limits, prior work~\cite{wu_sonata_2025, zhang_concerto_2025} suggests task-specific tuning could yield further gains. \project{} (M), despite not leveraging encoder pre-training and performing worst among variants, substantially increases t-mAP over the best non-temporal baseline, demonstrating the advantages of joint spatio-temporal refinement. Critically, our design avoids point-cloud superposition, preserving distinct instance identities across stages, while the bottlenecked sampling strategy enables efficient temporal integration with minimal memory overhead. Additional ablations and comparisons across strategies are discussed in \cref{sec:look}, with full results for Sonata and Minkowski backbones provided in Supp.

Table~\ref{tab:per_stage} reports per-stage 3DSIS mAP metrics, computed by evaluating the 4DSIS predictions independently at each stage in a sequence. This allows us to assess how well the standard 3DSIS task is preserved within the 4D setting. The experiment also highlights trade offs between different 3D matching strategies (Mask3D+sem, Mask3D+geo). While geometric matching yields higher overall performance when evaluated jointly (Table~\ref{tab:main_results}), it considerably degrades mAP between stages, due to not leveraging or fusing information from both scans. Notably, even though our method is not explicitly optimized for the 3D SIS task, it outperforms Mask3D, which is trained specifically for this setting. This demonstrates the benefit of sharing temporal information: by fusing multiple observations across scans, our model improves segmentation robustness and consolidates partial object views, effectively acting as a form of data and observation augmentation.

\begin{table}[t]
    \centering
    \footnotesize 
    \setlength{\tabcolsep}{4pt} 
    \begin{tabular}{lcccc}
      \toprule
      \textbf{Method}   & \textbf{Stage} & \textbf{mAP}  & \textbf{mAP$_{50}$}  & \textbf{mAP$_{25}$} \\
      \midrule
      Mask3D            & --   & \underline{46.4} & \textbf{68.5}  & \underline{78.5} \\
      + geo  & 2    & 21.9 & 46.4  & 68.4 \\
      \cmidrule(lr){1-5}
      Mask4Former       & 1    & 23.9 & 46.8  & 67.4 \\
                        & 2    & 23.2 & 48.0  & 67.6 \\
      \cmidrule(lr){1-5}
      \project{} (C)       & 1    & \textbf{47.8} & \underline{68.4} & \textbf{82.0} \\
                        & 2    & \textbf{48.3} & \textbf{69.8} & \textbf{83.0} \\
      \bottomrule
    \end{tabular}
    \caption{\textbf{Per-stage 3DSIS mAP on  3RScan \cite{wald_rio_2019}}. Joint 4D predictions are evaluated independently at each 3D temporal stage.}
    \label{tab:per_stage}
    \vspace{-12pt}
\end{table}

\subsection{Learning to Look Both Ways}
\label{sec:look}
We perform ablation studies to analyze the impact of temporal information sharing strategies on spatio-temporal instance segmentation performance.
To do this, we evaluate our three temporal sharing methods using our best performing backbone from Table~\ref{tab:main_results} (\project{} (C)): contrastive loss across time ($\mathcal{L}_{\mathrm{contr}}$), spatio-temporal serialization (ST-serial), and spatio-temporal masking (ST-mask), with results summarized in Table~\ref{tab:sharing_info}. All strategies individually or in combination improve sequence-level segmentation accuracy (t-mAP), with the most significant gains observed with a combination of contrastive loss and ST serialization. However, the effects saturate, suggesting a limitation in the total temporal signal. We attribute this to a dataset limitation: 3RScan has relatively few instances which change (17\% in validation) and inconsistent temporal instance annotations, mainly for foreground objects. Results for the other \project{} backbones are in Supp. We observe that the backbone determines which temporal sharing strategies yield the highest gains, reflecting differences in feature representation and latent space properties.

To understand the effect of each information sharing method, we report mean temporal recall (t-mREC) for instances categorized as static, rigid movement, non-rigid movement, and ambiguous; added/removed objects are omitted due to limited representation. We further provide t-REC for change types of interest:
\begin{itemize}
    \item \textit{Ambiguous instances}: Any form of temporal sharing helps disambiguate ambiguous instances. Although Concerto’s 2D–3D self-supervised features~\cite{zhang_concerto_2025} are semantically strong, using them without temporal context often causes visually similar objects to merge. Introducing temporal cues—via contrastive learning and temporal serialization—best mitigates this issue by adding discriminative signals from negative pairs within an ambiguous group and improved geometric alignment.
    \item \textit{Rigid changes}: For 4DSIS, \textbf{combining all temporal information strategies yields the best results on rigid changes.} While all methods improve performance, contrastive loss has the strongest individual effect by aligning features across time rather than relying solely on geometric cues. Although ST-mask and ST-serialization independently degrade rigid performance, together they are complementary. For instances with partial overlap, geometric cues still aid convergence, and ST-attention in the feature and query decoder help better distinguish features and queries of different instances.
    \item \textit{Non-rigid changes}: Deformable or shape-varying objects benefit most from ST-mask, which promotes spatial alignment across scans. This improves convergence for objects with feature deviation due to varying local geometry yet limited spatial movement, such as pillows and curtains.
    \item \textit{Mean t-recall}: Results depend on dataset composition, as common classes (e.g., chairs for ambiguous, curtains for non-rigid, and doors for rigid) tend to dominate specific change types. Because change types are unequally represented, overall t-mAP trends may vary accordingly.
\end{itemize}
Additional ablations focused on the architecture decision within temporal sharing, shown in Table~\ref{tab:st_ablation}. We demonstrate that mixing spatial and temporal serialization patterns further improves performance compared to spatial- or temporal-only serialization. Incorporating contrastive positive and negative pairs across time steps also boosts per-stage mAP, outperforming spatial-only contrastive loss and ST-joint query refinement.

\begin{table}[]
    \vspace{4pt}
    \small 
    \centering
\resizebox{0.98\columnwidth}{!}
{
    \begin{tabular}{@{}ccc|c|cccc@{}}
        \toprule
        $L_\mathrm{contr}$ & ST-serial & ST-mask & \textbf{t-mAP}  & \textbf{t-mREC}  & \textbf{Amb.} & \textbf{Rig.} & \textbf{Non-Rig.}\\
        \midrule
        \rowcolor{baseline}
        $\times$      & $\times$      & $\times$      & 28.4  & 41.8 & 20.4 & 44.9 & 62.1  \\
        $\checkmark$  & $\times$      & $\times$      & \underline{34.1}  &  49.6 & 42.8 & 48.4 &  63.2 \\
        $\times$      & $\checkmark$  & $\times$      & 32.9  &  48.8 & 43.2 & \cellcolor{palegray}40.9 & 67.0\\
        $\times$      & $\times$      & $\checkmark$  & 32.4  & 48.5 & 42.3 & \cellcolor{palegray}40.2 & \textbf{70.7} \\
        $\checkmark$  & $\checkmark$  & $\times$      & \cellcolor{good}\textbf{34.8}  &  52.1 & \underline{47.2} &  48.6 & \cellcolor{palegray}66.5 \\
        $\checkmark$  & $\times$      & $\checkmark$  & \cellcolor{palegray}33.2  &   \textbf{53.5} & \textbf{50.2} & \underline{54.4} & \cellcolor{palegray}65.7 \\
        $\times$      & $\checkmark$  & $\checkmark$  & \cellcolor{palegray}32.5  & \cellcolor{palegray}50.4 & \cellcolor{palegray}43.2 & 50.0 & \cellcolor{palegray}65.1 \\
        $\checkmark$  & $\checkmark$  & $\checkmark$  & \cellcolor{palegray}33.3  & \cellcolor{palegray}\underline{53.0} & \cellcolor{palegray}43.9 & \textbf{56.4} & \cellcolor{palegray}\underline{68.0}\\
        \bottomrule
    \end{tabular}
    }
    \caption{
    \textbf{Ablation of temporal information sharing strategies in 4DSIS.} 
    Rows show combinations of contrastive loss ($L_\mathrm{contr}$), spatio-temporal serialization (ST-serial), and spatio-temporal masking (ST-mask), with t-mAP and temporal recall (t-mREC) reported for ambiguous, rigid, and non-rigid instances. Best overall t-mAP is in \colorbox{good}{green}, baseline in \colorbox{baseline}{blue}, and \colorbox{palegray}{gray} denotes combinations that do not exceed individual method performance.
    }
    \label{tab:sharing_info}
\end{table}

\begin{table}
\centering
\vspace{-4pt}
\begin{minipage}{0.48\linewidth}
    \centering
    \footnotesize 
    \begin{tabular}{@{}ccl@{}}
        \toprule
        \multicolumn{1}{c}{\textbf{Pattern}} & \multicolumn{1}{c}{\textbf{t-mAP}} \\
        \midrule
        \Circled{1} 3D & 28.4 \\
        \Circled{2} 4D & 32.0 \\
        \Circled{3} 3D \& 4D & \textbf{32.9} \\
        \bottomrule
    \end{tabular}\\
    \textbf{a) Serialization}
\end{minipage}
\hfill
\begin{minipage}{0.48\linewidth}
    \centering
    \footnotesize
    \begin{tabular}{@{}ccc@{}}
        \toprule
        \multicolumn{1}{c}{\textbf{Scene}} & \multicolumn{1}{c}{\textbf{$\mathcal{L}_{\mathrm{contr}}$}} & \multicolumn{1}{c}{\textbf{mAP}} \\
        \midrule
        3D   & $\times$ &  42.3 \\
        3D   & $\checkmark$  &  43.6 \\
        4D   & $\times$ &  43.2\\
        4D   & $\checkmark$ &  \textbf{47.3}\\
        \bottomrule
    \end{tabular}\\
    \textbf{b) Contrastive Loss}
\end{minipage}
\caption{
\textbf{Temporal Information Sharing Ablations.} 
\textbf{a)} We study how different serialization patterns affect performance using the Concerto backbone: \Circled{1} Spatial Only (3D) \Circled{2} Temporal only (4D) where all serialization patterns traverse the 4D point cloud \Circled{3}, Spatio-temporal (3D \& 4D)
where we randomly shuffle temporal and spatial patterns. \textbf{b)} We evaluate the impact of using positive and negative temporal pairs across temporal stages in contrastive loss. Training and inference on either 3DSIS or 4DSIS. For 4D, mAP is averaged across stages for direct comparison.}
\label{tab:st_ablation}
\vspace{-10pt}
\end{table}

\section{Conclusion}
In this work, we formalize the task of 4D spatio-temporal semantic instance segmentation for indoor environments, propose a temporal evaluation metric tailored to this setting, and introduce our method, \project{}. By refining instance-level joint spatio-temporal queries and sharing information across temporal observations, our approach achieves consistent improvements in both segmentation quality and temporal association. Importantly, our design enables effective fusion of spatial and temporal information without relying on strict geometric assumptions, making it robust to real-world scenes with unobserved changes. 
While our findings highlight a promising direction for future research, progress is constrained by limited dataset diversity and annotation quality. This underscores the need for larger, more dynamic annotated datasets. Further limitations and future work directions are discussed in the Supp.

\paragraph{Acknowledgment}
This work is supported by the Stanford Institute for Human-Centered Artificial Intelligence (HAI), and E.S. is supported by the TomKat Center for Sustainable Energy as a TomKat Center Graduate Fellow for Translational Research. Stanford’s Marlowe computing clusters provided GPU computing for model training and evaluation.

{
    \small
    \bibliographystyle{ieeenat_fullname}
    \bibliography{references}
}

\clearpage
\setcounter{page}{1}
\maketitlesupplementary
\appendix
\renewcommand{\thesection}{\Alph{section}}


\newcommand{\temporalInstanceFigure}[5]{%
  \begin{figure*}[t]
    \centering
    \StrBehind{#4}{:}[\figbasename]%
    \includegraphics[width=\linewidth]{images/\figbasename}%
    \caption{#5}%
    \label{#4}%
  \end{figure*}%
}

\newcommand{\temporalInstanceFigureOurs}[5]{%
  \begin{figure*}[t]
    \centering
    \StrBehind{#4}{:}[\figbasename]%
    \includegraphics[width=\linewidth]{images/\figbasename}%
    \caption{#5}%
    \label{#4}%
  \end{figure*}%
}

\newcommand{\temporalInstanceFigureOursBeside}[5]{%
  \begin{figure*}[t]
    \centering
    \StrBehind{#4}{:}[\figbasename]%
    \includegraphics[width=\linewidth]{images/\figbasename}%
    \caption{#5}%
    \label{#4}%
  \end{figure*}%
}

\newcommand{\temporalInstanceFigureSharing}[5]{%
  \begin{figure*}[t]
    \centering
    \StrBehind{#4}{:}[\figbasename]%
    \includegraphics[width=\linewidth]{images/\figbasename}%
    \caption{#5}%
    \label{#4}%
  \end{figure*}%
}

\newcommand{\temporalJointFigure}[5]{%
  \begin{figure*}[t]
    \centering
    \StrBehind{#4}{:}[\figbasename]%
    \includegraphics[width=\linewidth]{images/\figbasename}%
    \caption{#5}%
    \label{#4}%
  \end{figure*}%
}

\newcommand{\temporalInstanceFigureScannet}[5]{%
  \begin{figure*}[t]
    \centering
    \StrBehind{#4}{:}[\figbasename]%
    \includegraphics[width=\linewidth]{images/\figbasename}%
    \caption{#5}%
    \label{#4}%
  \end{figure*}%
}

\noindent In this supplementary material we provide: 
\begin{itemize}
    \item A video discussing our method and results (Section \ref{supp:video})
    \item Visualizations of instance mask and semantic predictions (Section \ref{supp:qualitative})
    \item Additional experimental results and discussion (Section \ref{supp:ablations})
    \item Additional implementation details (Section \ref{supp:implementation})
    \item A discussion of the limitations of our method (Section \ref{supp:limitations}) 
\end{itemize}

\section{Video}
\label{supp:video}
Supplementary video available at \url{https://www.easteine.com/rescene4d/} summarizes the method, quantitative and qualitative results, and our temporal sharing ablation experiment.

\section{Qualitative Results}
\label{supp:qualitative}
We provide additional qualitative results showcasing instance masks and semantic predictions on several scenes from the 3RScan validation set \cite{wald_rio_2019}. Note that the 3RScan test set is not available for download, nor is there an evaluation server, so all results are reported on the validation split. For visualization, we project the point cloud predictions onto the original mesh vertices. 

We visualize the instance predictions from: Mask4Former~\cite{yilmaz_mask4former_2024}; Mask3D~\cite{schult_mask3d_2023} with post-hoc temporal semantic matching; and our best-performing model, \project{} (Concerto with ST-serialization and contrastive loss) in Figures~\ref{fig:main_example}, \ref{fig:close_up}, \ref{fig:baseline_problems}, \ref{fig:merging_stools}, \ref{fig:partial_overlap}, \ref{fig:ball}, \ref{fig:ambiguities_challenge}, \ref{fig:sharing}, \ref{fig:no_overlap}, \ref{fig:bathroom}, and \ref{fig:ambiguities_metric}. We omit Mask4D~\cite{marcuzzi_mask4d_2023} due to its poor performance, and we exclude Mask3D with geometric matching, as semantic matching provides a better balance between overall t-mAP and per-stage mAP. While geometric matching can reliably track well-aligned static instances, it struggles in sequences with partial overlap or instance changes, often degrading mask quality in the second stage. Note, instance colors are assigned independently for each method and do not correspond between methods or to ground truth; colors indicate distinct instances in instance segmentation, consistent across the temporal sequence to show shared identities.

For semantic segmentation, we visualize results using semantic category colors defined by ScanNet~\cite{dai_scannet_2017}. These colors are consistent across ground truth and all methods, ensuring direct visual comparability of semantic predictions. Semantic predictions are presented in Figures ~\ref{fig:merging_stools} and ~\ref{fig:partial_overlap}. Across methods, semantic segmentation performance remains comparable, with similar difficulties arising for challenging classes such as pillows and other small objects.
The challenge of the 3RScan dataset is evident in semantic predictions for all approaches, including ours, and in the degraded 3DSIS baseline mAP for single (non-temporal) 3D scans (see Figures ~\ref{fig:partial_overlap} and~\ref{fig:bathroom}). For example, Mask3D (without posthoc processing) achieves a validation mAP of 55.2 on ScanNet \cite{dai_scannet_2017}, but only 46.4 on 3RScan. This drop likely reflects the limited semantic diversity, increased noise, and smaller scale of 3RScan.

\temporalInstanceFigureOurs{scene0029}{00}{01}{fig:main_example}{%
  \textbf{Temporal Semantic Instance Segmentation Comparison.}
}
\temporalInstanceFigure{scene0029close}{00}{01}{fig:close_up}{%
  \textbf{Temporal Semantic Instance Segmentation Comparison---Close up of Figure~\ref{fig:main_example} with baseline comparison.} While Mask4Former has a tendency to merge multiple objects into one instance and Mask3D fails to track identities across time, \project{} consistently identifies and tracks object instances across temporal scans, be that dynamic or static.  It even considers the two bookcases next to the curtains as separate instances, in contrast to ground truth annotations. Non-systematic annotation of small objects throughout the dataset leads to our method missing some of the pillows on the couch, as well as the identification of a box on the top shelf of the bookcase as a separate instance. 
}

\temporalInstanceFigure{scene0269}{00}{02}{fig:baseline_problems}{%
    \textbf{Temporal Semantic Instance Segmentation Comparison—Common Baseline Errors.} Mask4Former maintains consistent identities across temporal stages but tends to merge objects---notably the couch and all pillows. Mask3D fails to track identities and has difficulty producing accurate instance masks in the first stage, where the couch is only partially observed. In contrast, \project{} correctly preserves identities despite bench movement and successfully segments most pillows, despite challenging non-rigid changes and partial observations.
}

\section{Additional Experimental Results }
\label{supp:ablations}
In this section, we further ablate the feature encoder and temporal sharing. 

\subsection{Temporal Sharing Experiments Details}
Table~\ref{tab:backbone_compare} compares t-mAP for Concerto (Conc.), Sonata (Son.), and Minkowski (Mink.) under different temporal sharing strategies. As discussed below, the optimal strategy differs across backbones. Table~\ref{tab:concerto_full}, ~\ref{tab:full_sonata}, and ~\ref{tab:full_mink} present the full ablation of each backbone reporting the t-mAP and t-REC (mean and per change type). In addition, to complement our core evaluation with t-mAP and t-REC, we report and analyze the mean inter-stage instance feature cosine similarity, or t-sim, across our experiments. t-sim measures, for each instance that is present in multiple temporal scans, the average similarity of its internal feature representation between paired stages. Unlike mAP or recall, which are defined by mask matches and ground truth, t-sim quantifies the temporal consistency of the learned instance embeddings independent of downstream query refinement or matching success.

\begin{table}[]
    \centering
    \small 
    \begin{tabular}{@{}ccc|c|c|c@{}}
        \toprule
        $L_\mathrm{contr}$ & ST-serial & ST-mask & \textbf{Conc.} & \textbf{Son.} & \textbf{Mink.} \\
        \midrule
        \rowcolor{baseline}
        $\times$      & $\times$      & $\times$      & 28.4  & 29.7  &  \textbf{32.0}   \\
        $\checkmark$  & $\times$      & $\times$      & 34.1  & 25.7  &  31.5  \\
        $\times$      & $\checkmark$  & $\times$      & 32.9  & 28.1  &   --  \\
        $\times$      & $\times$      & $\checkmark$  & 32.4  & 28.3  &  30.6   \\
        $\checkmark$  & $\checkmark$  & $\times$      & \cellcolor{good}\textbf{34.8}  & 29.6  &  --   \\
        $\checkmark$  & $\times$      & $\checkmark$  & 33.2  & 27.8  &  \cellcolor{good}\textbf{32.0} \\
        $\times$      & $\checkmark$  & $\checkmark$  & 32.5  & \cellcolor{good}\textbf{33.2}  &   --  \\
        $\checkmark$  & $\checkmark$  & $\checkmark$  & 33.3  & 26.2  &  --   \\
        \bottomrule
    \end{tabular}
    \caption{\textbf{Temporal instance sharing ablations with different encoders for t-mAP.} Different encoders benefit from different strategies, depending on their strengths---e.g., contrastive loss harms Sonata since the features are mainly geometric, and Minkowski does not allow serialization as a ConvNet. Best result per encoder in \colorbox{good}{\textbf{green}}, baseline in \colorbox{baseline}{blue}.}
    \label{tab:backbone_compare}
    \vspace{4pt}
\end{table}

\paragraph{Concerto.}
Table~\ref{tab:concerto_full} reports the full results of our Concerto backbone ablation study, discussed in Section \ref{sec:results} of the main paper, including additional change labels for static and added/removed instances. Figure~\ref{fig:sharing} provides a visual comparison for the base spatio-temporal architecture and our best performing version.
\begin{itemize}
    \item \textit{Static instances:} Temporal recall (t-REC) for static objects is strongly correlated with overall t-mAP, as static objects are much more prevalent in the dataset and thus dominate evaluation metrics. As with overall mAP, each temporal information sharing strategy improves convergence for static objects—whether by bringing feature representations closer together (contrastive loss), enabling joint feature decoding, or aligning spatio-temporal masks. However, combining multiple strategies leads to diminishing returns.
    \item \textit{Added and Removed instances}: We group added and removed objects together, as both represent instances present in one stage and absent in another. Due to very few such examples in the dataset (1.5\% of instances in the validation set), this metric is sensitive to noise and initialization and may lack enough signal for models to learn to identify removals effectively. Performance for added/removed is lower with the Concerto backbone compared to Sonata and Minkowski; improvements in static recall often come at the cost of added/removed recall (failure to produce temporal correspondences will lead to perfect removal prediction). 
\end{itemize}

\begin{table*}[]
    \vspace{4pt}
    \small
    \centering
{
    \begin{tabular}{@{}ccc|c|c|cccccccc@{}}
        \toprule
        $L_\mathrm{contr}$ & ST-serial & ST-mask & \textbf{t-mAP} & \textbf{t-Sim} & \textbf{t-mREC}  & \textbf{Amb.} & \textbf{Rig.} & \textbf{Non-Rig.} & \textbf{Static} & \textbf{Add. \& Rem.} \\
        \midrule
        \rowcolor{baseline}
        $\times$      & $\times$      & $\times$      & 28.4  & 0.8640 & 41.8 & 20.4 & 44.9 & 62.1 & 40.0  & 21.6\\
        $\checkmark$  & $\times$      & $\times$      & \underline{34.1}  & \underline{0.9160} & 49.6 & 42.8 & 48.4 &  63.2 & 43.9 & \cellcolor{palegray}18.8 \\
        $\times$      & $\checkmark$  & $\times$      & 32.9  & 0.8846 & 48.8 & 43.2 & \cellcolor{palegray}40.9 & 67.0 & \underline{44.0} & \cellcolor{palegray}15.7\\
        $\times$      & $\times$      & $\checkmark$  & 32.4  & 0.8797 & 48.5 & 42.3 & \cellcolor{palegray}40.2 & \textbf{70.7} & 40.8 & \cellcolor{palegray}17.9\\
        $\checkmark$  & $\checkmark$  & $\times$      & \cellcolor{good}\textbf{34.8}  & \textbf{0.9198} & 52.1 & \underline{47.2} &  48.6 & \cellcolor{palegray}66.5 & \textbf{46.1} & \underline{21.7} \\
        $\checkmark$  & $\times$      & $\checkmark$  & \cellcolor{palegray}33.2  & \cellcolor{palegray}0.9116 & \textbf{53.5} & \textbf{50.2} & \underline{54.4} & \cellcolor{palegray}65.7 & \cellcolor{palegray}43.6 & \textbf{26.5}\\
        $\times$      & $\checkmark$  & $\checkmark$  & \cellcolor{palegray}32.5  & 0.8980 & 50.4 & \cellcolor{palegray}43.2 & 50.0 & \cellcolor{palegray}65.1 & \cellcolor{palegray}43.1 & \cellcolor{palegray}17.5\\
        $\checkmark$  & $\checkmark$  & $\checkmark$  & \cellcolor{palegray}33.3  & \textbf{0.9198} & \cellcolor{palegray}\underline{53.0} & \cellcolor{palegray}43.9 & \textbf{56.4} & \cellcolor{palegray}\underline{68.0} & \cellcolor{palegray}43.7 & \cellcolor{palegray}16.2 \\
        \bottomrule
    \end{tabular}
    }
    \caption{
    \textbf{Concerto Ablation of temporal information sharing strategies in 4DSIS.} 
    Rows show combinations of contrastive loss ($L_\mathrm{contr}$), spatio-temporal serialization (ST-serial), and spatio-temporal masking (ST-mask), with t-mAP and temporal recall (t-mREC) reported for ambiguous, rigid, non-rigid, static, and added/removed instances. Best overall t-mAP is in \colorbox{good}{green}, best per metric in \textbf{bold}, second best per metric \underline{underlined}, baseline in \colorbox{baseline}{blue}, and \colorbox{palegray}{gray} denotes combinations that do not exceed individual method performance.
    }
    \label{tab:concerto_full}
\end{table*}

\paragraph{Sonata.}
Table~\ref{tab:full_sonata} presents the full results of our Sonata backbone ablation study. Sonata achieves its highest t-mAP, t-mREC, and per-change-type performance when spatio-temporal masking is combined with spatio-temporal serialization. Serialization enables joint decoding of features across time, improving categories such as static and ambiguous. Masking alone enforces temporal consistency in instance masks during query refinement but does not explicitly promote feature alignment between temporal stages. When paired with spatio-temporal serialization, this feature alignment mechanism leads to more comprehensive improvements across change types.

\begin{figure}[t]
  \centering
  \begin{subfigure}{0.48\linewidth}
    \includegraphics[width=\linewidth]{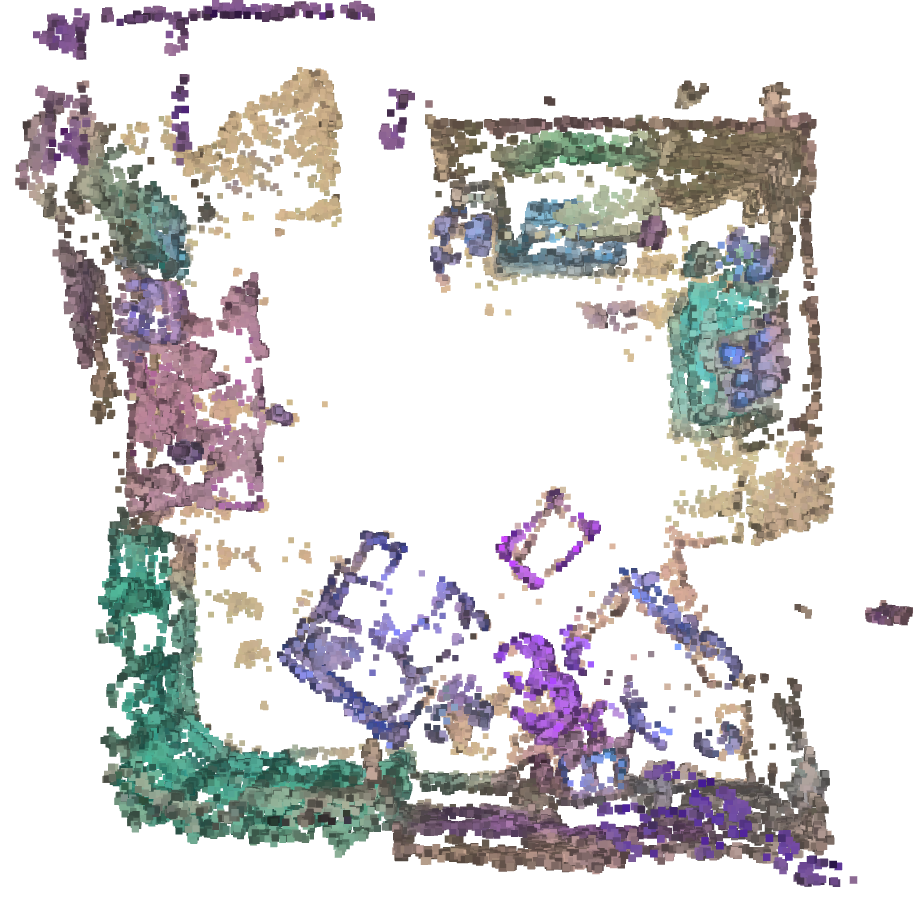}
    \label{fig:pca_concerto}
  \end{subfigure}
  \hfill
  \begin{subfigure}{0.48\linewidth}
    \includegraphics[width=\linewidth]{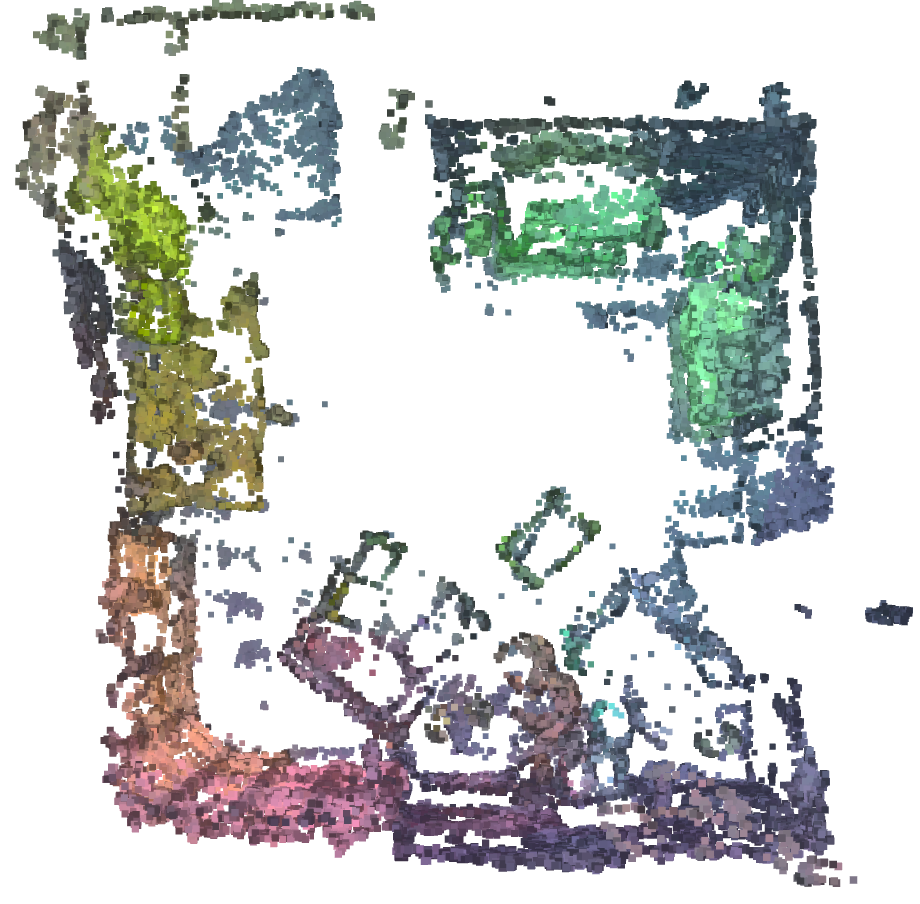}
    \label{fig:pca_sonata}
  \end{subfigure}
  \caption{\textbf{Feature PCA for a sample scene per PTv3 encoder.} Left — \textit{Concerto}, Right — \textit{Sonata}.}
  \label{fig:pca}
\end{figure}

In contrast to Concerto, contrastive loss significantly reduces performance in Sonata ablations. This emphasizes the dependence of temporal sharing strategies on backbone feature representations. Figure~\ref{fig:pca} illustrates this: a PCA visualization of point cloud features across the two different backbones (without temporal sharing modifications) shows Sonata features are primarily geometry-driven, and sensitive to global location. For large instances, substantial variation in point features within the same object may obscure the contrastive learning signal, potentially hindering effective temporal alignment. By contrast, Concerto generates more semantically oriented features. In this setting, contrastive loss aids discrimination between visually or semantically similar objects (e.g., ambiguous classes or varied instances such as different types of pillows), resulting in clear feature separation across time. Notably, with naturally strong geometric consistency, the Sonata canonical has better performance over the Concerto baseline, as its geometry-based features are stable across temporal stages and well-separated among instances within a scene. However, Concerto benefits more from temporal sharing and ultimately shows larger relative gains when these strategies are applied.

\begin{table*}[]
    \vspace{4pt}
    \small 
    \centering
{
\begin{tabular}{@{}ccc|c|c|cccccccc@{}}
    \toprule
    $L_\mathrm{contr}$ & ST-serial & ST-mask & \textbf{t-mAP} & \textbf{t-Sim} & \textbf{t-mREC}  & \textbf{Amb.} & \textbf{Rig.} & \textbf{Non-Rig.} & \textbf{Static} & \textbf{Add. \& Rem.} \\
    \midrule
    \rowcolor{baseline}
    $\times$      & $\times$      & $\times$      & 29.7 & 0.8053 & 45.3 & 30.9 & \underline{49.2} & \underline{58.9} & 42.3 & \underline{33.4}\\
    $\checkmark$  & $\times$      & $\times$      & \cellcolor{palegray}25.7 & 0.8767 & \cellcolor{palegray}42.1 & \cellcolor{palegray}30.0 & \cellcolor{palegray}43.7 & \cellcolor{palegray}55.6 & \cellcolor{palegray}39.3 & \cellcolor{palegray}21.3\\
    $\times$      & $\checkmark$  & $\times$      & \cellcolor{palegray}28.1 & \textbf{0.8949} & \underline{46.1} & \underline{39.3} & \cellcolor{palegray}49.0 & \cellcolor{palegray}53.4 & \underline{42.5} & \cellcolor{palegray}26.3\\
    $\times$      & $\times$      & $\checkmark$  & \cellcolor{palegray}28.3 & 0.8713 & \cellcolor{palegray}44.6 & \cellcolor{palegray}26.4 & 52.9 & \cellcolor{palegray}57.8 & \cellcolor{palegray}41.3 & \cellcolor{palegray}28.2\\
    $\checkmark$  & $\checkmark$  & $\times$      & \cellcolor{palegray}29.6 & \cellcolor{palegray}0.8811 & \cellcolor{palegray}44.8 & \cellcolor{palegray}32.4 & \cellcolor{palegray}48.6 & \cellcolor{palegray}56.8 & \cellcolor{palegray}41.5 & \cellcolor{palegray}31.5\\
    $\checkmark$  & $\times$      & $\checkmark$  & \cellcolor{palegray}27.8 & \cellcolor{palegray}0.8708 & \cellcolor{palegray}42.3 & 31.5 & \cellcolor{palegray}52.1 & \cellcolor{palegray}48.0 & \cellcolor{palegray}39.8 & \cellcolor{palegray}26.4\\
    $\times$      & $\checkmark$  & $\checkmark$  & \cellcolor{good}\textbf{33.2} & \cellcolor{palegray}\underline{0.8885} & \textbf{52.3} & \textbf{44.6} & \textbf{56.1} & \textbf{65.2} & \textbf{43.2} & \cellcolor{palegray}29.0\\
    $\checkmark$  & $\checkmark$  & $\checkmark$  & \cellcolor{palegray}26.2 & \cellcolor{palegray}0.8791 & \cellcolor{palegray}39.5 & \cellcolor{palegray}20.7 & \cellcolor{palegray}51.0 & \cellcolor{palegray}48.8 & \cellcolor{palegray}37.6 & \textbf{35.7}\\
    \bottomrule
\end{tabular}
    }
    \caption{
    \textbf{Sonata Ablation of temporal information sharing strategies in 4DSIS.} 
    Rows show combinations of contrastive loss ($L_\mathrm{contr}$), spatio-temporal serialization (ST-serial), and spatio-temporal masking (ST-mask), with t-mAP and temporal recall (t-mREC) reported for ambiguous, rigid, non-rigid, static, and added/removed instances. Best overall t-mAP is in \colorbox{good}{green}, best per metric in \textbf{bold}, second best per metric \underline{underlined}, baseline in \colorbox{baseline}{blue}, and \colorbox{palegray}{gray} denotes combinations that do not exceed individual method performance.
    }
    \label{tab:full_sonata}
\end{table*}

\paragraph{Minkowski.}
Table~\ref{tab:full_mink} presents the complete results of our Minkowski backbone ablations. Minkowski shows the least improvement over its canonical model—largely because its canonical version outperforms those of other backbones. This is expected: it is the only encoder trained directly on this task. However, training the backbone from scratch prioritizes learning 3D feature representations and adapting to the dataset distribution, resulting in most improvements stemming from spatial learning rather than effective use of temporal information.  The key takeaway is that joint spatio-temporal query refinement—central to all of our approaches—provides a substantial advantage over existing models such as Mask3D and Mask4Former.

Individually, both contrastive loss and spatio-temporal masking moderately enhance performance for certain change types, particularly in t-mREC. When combined, these strategies yield gains for ambiguous, rigid, and non-rigid categories compared to the canonical model. With full temporal information sharing, Minkowski maintains overall t-mAP while improving recall across change types, which better reflects performance relative to the frequency of changes in the dataset.

\begin{table*}[]
    \vspace{4pt}
    \small 
    \centering
{
    \begin{tabular}{@{}ccc|c|c|ccccccc@{}}
        \toprule
        $L_\mathrm{contr}$ & ST-mask & \textbf{t-mAP} & \textbf{t-Sim} & \textbf{t-mREC} & \textbf{Amb.} & \textbf{Rig.} & \textbf{Non-Rig.} & \textbf{Static}  & \textbf{Add. \& Rem.} \\
        \midrule
        \rowcolor{baseline}
        $\times$      & $\times$      & \textbf{32.0} & 0.9228 & 44.9 & 44.6 & 49.2 & 41.6 & \underline{44.9} & \underline{29.8}\\
        $\checkmark$  & $\times$      & \cellcolor{palegray}31.5 & \underline{0.9251} & \underline{47.3} & \cellcolor{palegray}43.5 & \textbf{56.0} & \underline{44.8} & \textbf{45.6} & \textbf{32.3}\\
        $\times$      & $\checkmark$  & \cellcolor{palegray}30.6 & 0.9238 & \textbf{48.6} & \textbf{49.4} & \underline{54.4} & \textbf{48.0} & \cellcolor{palegray}44.2 & \cellcolor{palegray}28.5 \\ 
        $\checkmark$  & $\checkmark$  & \cellcolor{good}\textbf{32.0} & \textbf{0.9261} & \cellcolor{palegray}47.2 & \underline{48.9} & \cellcolor{palegray}52.9 & \cellcolor{palegray}44.5 & \cellcolor{palegray}44.2 & \cellcolor{palegray}29.0  \\
        \bottomrule
    \end{tabular}
    }
    \caption{
    \textbf{Minkowski Ablation of temporal information sharing strategies in 4DSIS.} 
    Rows show combinations of contrastive loss ($L_\mathrm{contr}$), spatio-temporal serialization (ST-serial), and spatio-temporal masking (ST-mask), with t-mAP and temporal recall (t-mREC) reported for ambiguous, rigid, non-rigid, static, and added/removed instances. Best overall t-mAP is in \colorbox{good}{green}, best per metric in \textbf{bold}, second best per metric \underline{underlined}, baseline in \colorbox{baseline}{blue}, and \colorbox{palegray}{gray} denotes combinations that do not exceed individual method performance.
    }
    \label{tab:full_mink}
\end{table*}

\paragraph{Analysis of the t-Sim Metric.}
As expected, temporal information sharing strategies consistently improve t-sim over the canonical backbone for all encoders (see Tables~\ref{tab:concerto_full}, \ref{tab:full_sonata}, \ref{tab:full_mink}), though the magnitude and baseline differ:

\begin{itemize}
    \item \textbf{Concerto:} Displays a substantial increase in t-sim when temporal sharing mechanisms are used, with the best temporal consistency from the combination of contrastive loss and serialization. These settings also correspond to strong t-REC and recall for rigid instances.
    \item \textbf{Sonata:} Shows the largest improvement in t-sim between base and sharing variants. For Sonata, t-sim most clearly identifies serialization as the best mechanism for temporal sharing with respect to temporal feature consistency.
    \item \textbf{Minkowski:} For Minkowski, temporal sharing mechanisms yield smaller absolute gains in t-sim compared to the other backbones but remain beneficial, especially when both contrastive loss and temporal masking are employed. The base Minkowski model, trained directly on the 4DSIS 3RScan task and without explicit temporal information sharing, already achieves a high t-sim (0.92). In contrast, Concerto and Sonata, as pre-trained encoders not specifically designed for 4DSIS, show lower baseline inter-stage consistency.
\end{itemize}

While t-sim is a useful indicator of temporal feature consistency, it does not perfectly track segmentation performance. The highest t-sim configurations nearly always achieve high t-mAP, but not always the overall best. However, t-sim aligns more closely with t-REC, where high temporal consistency is important for instance recovery across change types. Notably, Minkowski achieves the highest absolute t-sim but not the best t-mAP or t-REC. This suggests that while highly consistent features are necessary for robust tracking, segmentation quality also hinges on effective query refinement and mask generation. Overall, t-sim provides a valuable auxiliary diagnostic for assessing temporal consistency in feature space, complementing mask-based metrics and revealing the encoder-dependent effects of explicit temporal sharing.

\subsection{Longer Sequence Lengths}
\begin{figure}[h!]
    \vspace{-10pt}
    \centering
        \includegraphics[width=\linewidth]{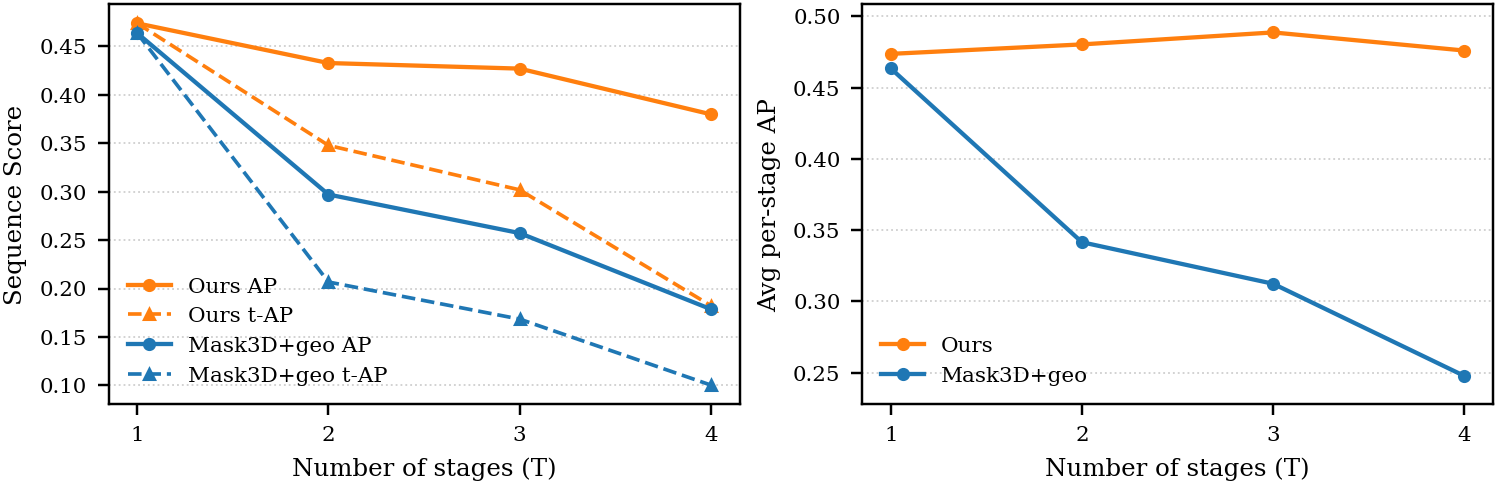}
        \captionsetup{font=scriptsize}
        \caption{4DSIS and Per-Stage 3DSIS evaluation metrics for \project{} inference on sequences of varying lengths. }
        \label{fig:longer_sequences}
\end{figure}

We show preliminary results for sequences \(T>2\) with our model (inference only) outperforming Mask3D+geo in Figure~\ref{fig:longer_sequences}. When \(T>2\), identity drift is more challenging. Our metric t-mAP reflects this degradation, while per stage 3DSIS AP remains robust. This experiment demonstrates limitations in our model and suggests better long-term tracking will require training with more data and longer sequences. 

\subsection{Analysis of Baseline Performance}
\paragraph{Mask3D and Post-hoc Matching.}
Semantic and geometric matching in Mask3D struggle with partial overlaps and significant scene changes (see Figure~\ref{fig:partial_overlap}). Semantic matching is restricted to instances within the same predicted semantic class, so matches fail if corresponding objects are classified differently across scans—especially problematic in cases with partial observations. Meanwhile, geometric matching relies on simple nearest-neighbor assignments, which quickly breaks down as scene dynamics increase. Combining the two would require manually and potentially per-sequence tuned heuristics to become reliable. In contrast, our joint instance query refinement and direct spatio-temporal mask prediction robustly handle evolving indoor scenes without manual tuning. 

\paragraph{Mask4Former.}
We carefully examined Mask4Former ~\cite{yilmaz_mask4former_2024} due to its architectural resemblance to our approach, notably its joint prediction of spatio-temporal instance queries and masks adapted from Mask3D \cite{schult_mask3d_2023}. As discussed in Section~\ref{sec:results}, Mask4Former was developed for outdoor LiDAR-based 4D panoptic segmentation, and is configured to operate only on input sequences ($T > 1$); adapting it for single-frame training would require substantial pipeline changes. To maintain a faithful and reproducible baseline, we train Mask4Former only on 3RScan sequence data and did not train it with a mixed 3RScan + ScanNet set. To account for the advantages of including ScanNet, we adopted a transfer learning setup for Mask4Former: it was initialized with a ScanNet-pretrained Minkowski backbone (from Mask3D) and finetuned on 3RScan. To completely isolate the impact of pretraining compared to dataset mixing, we also trained our method  with the same backbone (\project{} M) under equivalent transfer learning training settings, where it outperforms Mask4Former by +14.9.

Mask4Former's core design choice—superimposing point clouds from multiple scans—undermines per-stage performance in temporally sparse indoor environments. This approach yields strong instance identity consistency across scans, demonstrated by the minimal drop from single- to multi-stage metrics (see Tables~\ref{tab:main_results} and \ref{tab:per_stage} in the main paper, and Figures ~\ref{fig:close_up}, ~\ref{fig:baseline_problems}, ~\ref{fig:merging_stools}, and ~\ref{fig:ball} for qualitative results). However, it incorporates temporal information (multiple scans) without an explicit temporal signal, severely degrading per-stage performance. Lacking differentiation between temporal stages, Mask4Former cannot distinguish object movement from duplication—for example, if a chair moves slightly to the left, it may interpret this as two chairs next to each other, or, if a pair of bowls is set on a table, it may merge their identities assuming that duplication means motion instead of recognizing them as separate objects. This issue is especially pronounced for small objects. The confusing training signal from overlaid scans results in frequent instance merging and failed tracking of moved objects within and across scans. Additionally, if an object appears in a spatial location where another object was present in the previous stag, Mask4Former’s architecture prevents assigning different identities or masks to the overlapping region (see Figure~\ref{fig:ball}).

Furthermore, its box loss enforces overly compact masks, which is ill-suited for objects with elongated shapes, such as doors or shelving units, leading to poor instance grouping in such cases. Since Mask4Former is also unable to train simultaneously with sequences and single-stage scans, we pretrain its Minkowski backbone using Mask3D on single-stage scans from ScanNet before training with the smaller 3RScan dataset. In contrast, our method can train directly on the mix of both datasets, hence allowing larger flexibility and an enhanced learning from both spatial and temporal signals. 

\paragraph{Mask4D.}
Similar to Mask4Former, the original training codebase of Mask4D~\cite{marcuzzi_mask4d_2023} is architecturally designed for sequential inputs and expects $T > 1$; however, its backbone architecture does not have existing ScanNet-pretrained single-scan models like Mask3D. Consequently, we trained the entire model, including its backbone, from scratch on the 3RScan dataset. Mask4D failed to converge reliably (see Table~\ref{tab:main_results}). We observed that Mask4D propagates a very limited number of instance queries, further restricting its capacity to reliably track objects over time. To ensure a fair comparison, we also trained our method (\project{} M) under equivalent training settings (3RScan only, from scratch), where it outperforms Mask4D by +19.0.

\section{Additional Implementation Details }
\label{supp:implementation}
\subsection{Mask3D Background}
Additional Mask3D background is included for clarity and completeness of the description of our overall system \cite{schult_mask3d_2023}. These descriptions do not reflect our contributions. 
\label{supp:background}
\paragraph{Mask Module.} The mask module assigns each instance query $X \in \mathbb{R}^{K \times D}$ a binary mask and semantic class. Instance queries are mapped through an MLP $f_{\mathrm{mask}}(\cdot)$ to align with the backbone's fine-grained features $F_0 \in \mathbb{R}^{M_0 \times D}$, accumulated as means over superpoints. The binary mask $B \in \{0, 1\}^{M_0 \times K}$ is obtained via dot product, followed by sigmoid activation and thresholding \(B = \left[\sigma\left(F_0 f_{\mathrm{mask}}(X)^\top\right) > 0.5\right]\). For each query, semantic class probabilities are predicted via a linear projection onto $C + 1$ classes and softmax normalization. 

\paragraph{Query Refinement.} Instance queries $X \in \mathbb{R}^{K \times D}$ are initialized in a non-parametric fashion and progressively refined via a stack of transformer-based query decoder layers. Each layer alternates masked cross-attention to backbone features at multiple resolutions and self-attention between queries to mitigate query competition~\cite{schult_mask3d_2023}. At each resolution $r$, backbone voxel features $F_r \in \mathbb{R}^{M_r \times D_r}$ are linearly projected to keys $K$ and values $V$, while queries $X$ yield queries $Q$. Each query attends only to foreground voxels from a mask module produced mask prediction from the full resolution super point features and the previous layer queries, pooling masks to match the feature hierarchy. 

\paragraph{Loss.}
Mask3D supervises mask and semantic predictions using standard bipartite matching via the Hungarian algorithm, as in~\cite{schult_mask3d_2023}. The assignment cost combines Dice and binary cross-entropy mask losses, and multi-class semantic cross-entropy:
\[
C(k, k') = \lambda_{\text{dice}} \, \mathcal{L}_{\text{dice}} + \lambda_{\text{BCE}} \, \mathcal{L}_{\text{BCE}} + \lambda_{\text{cls}} \, \mathcal{L}_{\text{cls}}
\]
where $C(k, \hat{k})$ is the cost of assigning predicted instance $k$ to ground-truth instance $\hat{k}$; $\mathcal{L}_{\text{dice}}$ is the Dice loss; $\mathcal{L}_{\text{BCE}}$ is the binary cross-entropy loss over the mask; and $\mathcal{L}_{\text{cls}}$ is the multi-class cross-entropy classification loss. The weights $\lambda_{\text{dice}}$, $\lambda_{\text{BCE}}$, and $\lambda_{\text{cls}}$ balance the respective contributions of each loss component, and are set to $\lambda_{\text{dice}} = \lambda_{\text{cls}} = 2.0$, $\lambda_{\text{BCE}} = 5.0$.

\subsection{\project{} Method Details}
\paragraph{Defining 4D Spatio-Temporal Sequences.}
For each scene \textit{sequence}, we annotate temporal changes by comparing instance labels and transformations between scans, labeling object ambiguities, non-rigid deformations, rigid movements, and added/removed instances. These per-vertex change labels are used for downstream analysis. Since these sequences have arbitrary lengths between measurements, we populate the time coordinate of the 4D point clouds with the index of each stage. 

\paragraph{Contrastive Loss.}
InfoNCE loss traditionally employs cosine similarity scaled by a temperature parameter. To avoid dataset-dependent temperature tuning or learnable temperature, we adopt the temperature-free log-odds normalization~\cite{kim_temperature-free_2025}. Given superpoint features $f_i, f_j \in \mathbb{R}^D$, the log-odds normalized cosine similarity matrix $L \in \mathbb{R}^{S \times S}$ is:
\begin{equation}
   L_{ij} = 2 \cdot \mathrm{atanh}\left( \frac{f_i^\top f_j}{\lVert f_i \rVert \lVert f_j \rVert} \right)
    \label{eq:logodds_dotprod}
\end{equation}
where $f_i^\top f_j / (\lVert f_i \rVert \lVert f_j \rVert)$ is the cosine similarity between superpoints $i$ and $j$.

\paragraph{Training Implementation.} Unless otherwise specified, we adopt the hyperparameter settings from Mask3D~\cite{schult_mask3d_2023}, including 3D data augmentations such as random rotation and scaling, which are applied to entire registered sequences together. For all experiments, we use $N_q=100$ queries and set the empty object loss weighting to $\lambda_{\text{empty}}=0.2$, regardless of temporal sequence length. The mixed [3RScan:ScanNet] training ratio was selected based on preliminary Minkowski(M) backbone experiments for reasonable performance. Query initialization is performed via Farthest Point Sampling (FPS) on input point positions. Spatio-temporal point clouds are voxelized at a resolution of 2~cm. Models are trained for 450 epochs with a batch size of 32, using the AdamW optimizer and a one-cycle learning rate scheduler, with a maximum learning rate of $5 \times 10^{-4}$. During training, the PTv3 pre-trained encoders, Sonata and Concerto, are frozen while their decoder is trained from scratch. For PTv3 backbones, we train across 8 NVIDIA H100 GPUs for 26 hours. For Minkowski backbone experiments we train across 2 NVIDIA H100 GPUs for 42 hours. We utilize Stanford's Marlowe computing cluster \cite{kapfer_marlowe_2025} for model training and evaluation. 

\subsection{Metric (t-mAP) Details}
\label{supp:metric}
We formalize additional details of our temporal mean Average Precision (\textbf{t-mAP}) and its treatment of ambiguous ground truth instances.

\paragraph{Matching.}
Similar to mAP, prior to thresholding, a prediction \( pr_i(c) \) is considered \emph{matched} to a ground truth instance \( gt_j(c) \) if
\(
\mathrm{IoU}(pr_i, gt_j) > 0,
\)
where both \( pr_i \) and \( gt_j \) belong to the same semantic class \( c \).
Each ground truth instance \( gt_j(c) \) may be matched by multiple predictions \( pr_i(c) \), forming the set of overlapping predictions for that ground truth:
\[
\mathcal{P}(gt_j(c)) = \left\{ pr_i \in pr(c)\ :\ \mathrm{IoU}(pr_i, gt_j(c)) > 0 \right\}.
\]

\paragraph{Disambiguating.}
For t-AP, we also match predictions to ambiguously similar sets of ground truth instances.  Let \(\mathcal{G}\) be an ambiguous group containing \(n_{\mathrm{amb}}\) ground truth instances over \(T\) stages. At each stage \(t\), there are up to \(n_{amb}\) ground truth instances resulting in \(\mathcal{G} := \{gt_k(t) \mid k \in [1, n_{\mathrm{amb}}], t \in [1, T]\}\). Let \(\mathcal{P}(\mathcal{G})\) be the set of predictions with overlap to any member \(gt_k(t)\) of \(\mathcal{G}\). The objective is to pseudo-disambiguate each ambiguous group by finding a set of unique trajectories \(\{\mathcal{G}'_i : i \in [n_{amb}]\}\)—each associating one ground truth instance per stage—guided by the predictions to prevent unfair penalization of symmetric identity swaps while penalizing merges. Figure~\ref{fig:ambiguities_metric} illustrates instances of identity switches that should not be penalized by the evaluation metric.

For each \(pr \in \mathcal{P(\mathcal{G})}\), \(k \in [1, n_{\mathrm{amb}}]\), and temporal stage \(t \in [1, T]\), we define the assignment weight as:
\[
W_{p,k,t} := \mathrm{IoU}(pr(t), gt_k(t)) \cdot \mathcal{C}_{pr(t)}
\]
where \(\mathcal{C}_{pr(t)}\) is the confidence score of prediction \(pr\). This weight quantifies how well each prediction supports a particular ground truth instance at each temporal stage, taking both overlap and prediction confidence into account.

The assignment process proceeds iteratively: at each step, the prediction with the highest total weight—i.e., exhibiting the best combination of overlap and confidence with a specific trajectory of ground truth instances—is selected and assigned to that trajectory. Subsequent predictions are assigned to the next-best groupings in a greedy fashion.

\begin{figure}[t]
  \centering
    \vspace{0pt} 
    \centering
    \includegraphics[width=\linewidth]{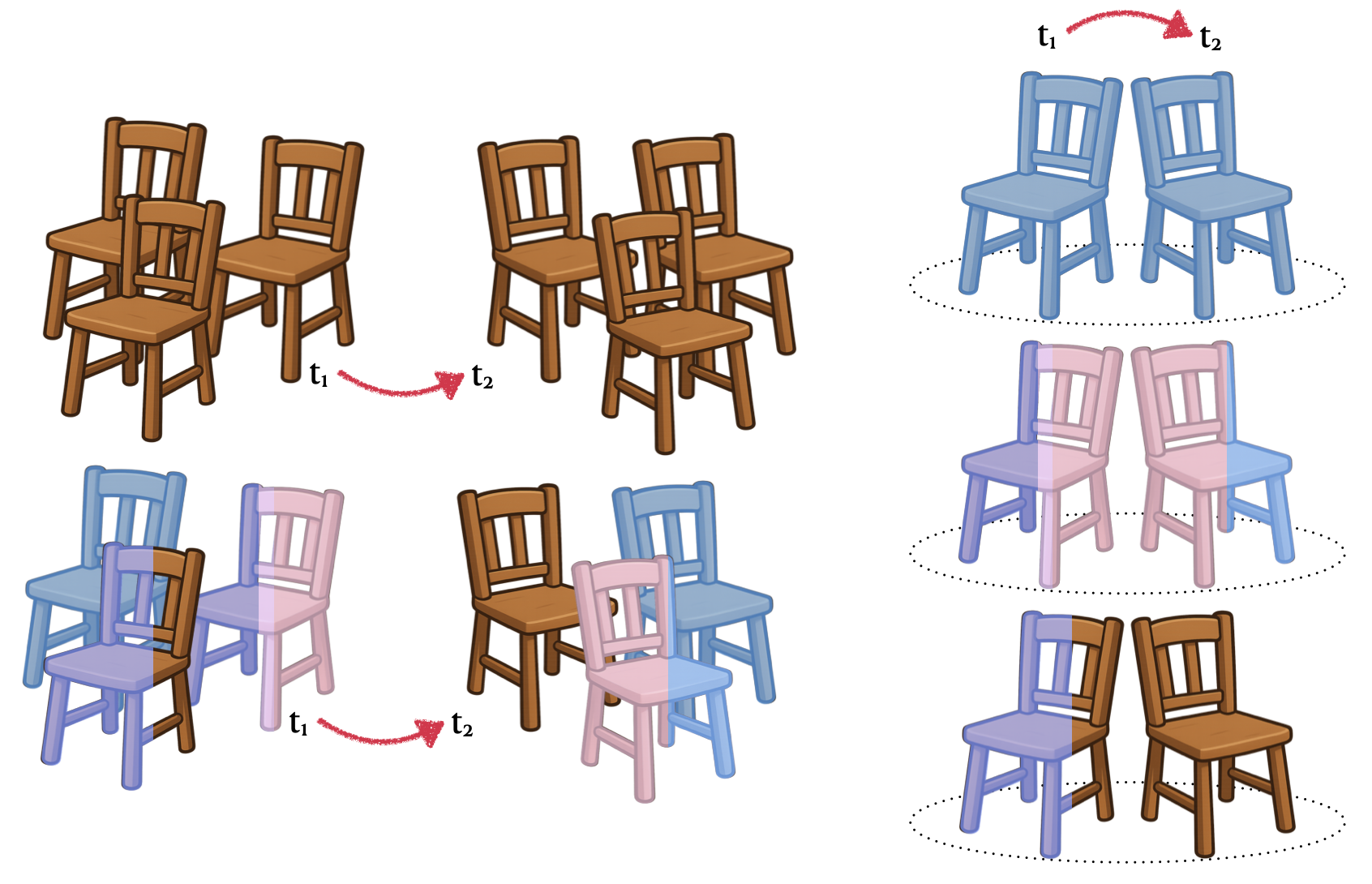}
  \caption{\textbf{Toy Example for Pseudo-Disambiguating Ambiguous Instance Groups.} Predictions are used to guide the process of creating disambiguated groups despite prediction conflicts or errors by ordering priority based on mask quality, overlap, and confidence}
  \vspace{-10pt}
  \label{fig:amb_toy}
\end{figure}

If a prediction does not overlap with a temporal stage, these stages in the trajectory are left unassigned. After all predictions have been exhausted or all k ground truth disambiguated trajectories are at least partially defined, any remaining unassigned ground truth instances are randomly assigned to available trajectories to ensure that each trajectory is completed. This process is defined in Algorithm~\ref{alg:assign_ambiguous} a toy visual example is included in Figure~\ref{fig:amb_toy}.

\begin{algorithm}[tb]
   \caption{Ambiguous Instance Assignment for t-mAP}
   \label{alg:assign_ambiguous}
   \begin{algorithmic}
      \STATE {\bf Input:} $\mathcal{G} = \{gt_k(t)\}$, $\mathcal{P}(\mathcal{G})$
      \STATE $W_{pr, k, t} \gets \mathrm{IoU}(pr(t), gt_k(t)) \cdot \mathcal{C}_{pr(t)}$\ \hfill$\forall\ pr \in \mathcal{P}(\mathcal{G}),\ k \in [1, n_{\mathrm{amb}}],\ t \in [1, T]$
      \STATE Initialize $A \in \mathbb{Z}^{n_{\mathrm{amb}} \times T} \gets \textit{unassigned}$
      \FOR{$i = 1,\ldots, n_{\mathrm{amb}}$}
         \STATE $p^* \gets \displaystyle \arg\max_{pr} \sum_{t} \max_{k}W_{pr, k, t}$
         \FOR{$t = 1,\ldots, T$}
            \STATE $k^* \gets \displaystyle \arg\max_{k} W_{p^*, k, t}$
            \IF{$W_{pr^*, k^*, t} > 0$}
               \STATE $A_{i, t} \gets k^*$
               \STATE $W_{k^*, t} \gets 0$
               \STATE $W_{p^*, t} \gets 0$
            \ENDIF
         \ENDFOR
      \ENDFOR
      \STATE Fill any $A_{k, t} = $\textit{unassigned} with available $gt_k(t)$
      \STATE Partition $\mathcal{G}$ into $\{\mathcal{G}'_i: i \in [1, n_{\mathrm{amb}}]\}$ using $A$ indices
      \RETURN Pseudo gt and matches $\{(\mathcal{G}'_i, \mathcal{P}(\mathcal{G}'_i))\}$
   \end{algorithmic}
\end{algorithm}

\paragraph{Assigning.}
For each class \(c\) and threshold \(\tau\), the sets of true positives, false positives, and false negatives are formulated from the overall set of predictions \(pr(c)\) and ground truth instances \(gt(c)\). Pseudo-disambiguated ground truth instances are treated as normal. Predictions are processed in order of descending confidence. A predicted instance \(pr_i \in pr(c)\) is assigned as a true positive, \(TP_c\), to a ground truth \(gt_j \in gt(c)\) if their temporal intersection-over-union (\(\mathrm{t}\text{-}\mathrm{IoU} > \tau\)), and \(gt_j\) has not already been assigned. For each prediction, among all eligible ground truth instances, assignment is made to the \(gt_j\) with the highest overlap. Each ground truth instance receives at most one assigned prediction, and each prediction is assigned to at most one ground truth instance. False positives, \(FP_c\), and False negatives, \(FN_c\), are defined by the set of unpaired predictions and ground truth.

Once \(TP_c\), \(FP_c\), and \(FN_c\) are determined, precision-recall curves are computed using prediction confidences to calculate \(t\)-AP for each class; \(t\)-mAP is their mean. Additionally, we report per-change-type temporal recall by grouping true and false negatives by annotated instance change and aggregating recall across all classes.

We create a custom PyTorch Lightning Metric for our t-mAP and t-mREC which will be released publicly.

\section{Limitations}
\label{supp:limitations}

Our experiments on 3RScan and ScanNet demonstrate improved performance and robustness over baseline methods, highlighting the importance of explicit temporal information sharing and unified architecture for joint predictions. However, the effectiveness of temporal modeling is closely linked to the dataset, with current benchmarks such as 3RScan limited by semantic class diversity and the distribution of changes (dominated by static instances). 
Despite these data limitations, methods with strict geometric assumptions (Mask3D+geo, Mask4Former) still do not outperform ours, even in scenes with minimal changes. This demonstrates that, even under conditions that ideally favor geometric methods, our approach remains more robust to the unique challenges of temporally sparse 4D indoor segmentation, such as partial scans, noise, and small misalignment. Though good registration is generally achievable with modern methods, different training strategies for non-registered scans could be included in future work. Progress in this area urgently requires more diverse, large-scale annotated 4D indoor datasets of changing scenes to better support and evaluate advanced and realistic 4D tasks with more significant geometric variation.

Computational complexity remains a challenge; the use of temporally paired scans increases memory and processing demands, restricting our experiments to short temporal windows and moderate scene sizes.

Additionally, our focus was to demonstrate the impact of temporal information sharing, and we did not perform extensive hyperparameter tuning or incorporate the latest 3DSIS query and feature refinement approaches presented in recent literature~\cite{lu_query_2023, wang_competitorformer_2025, lai_mask-attention-free_2023, lu2025relation3d}. Future work could integrate these advances to further boost prediction accuracy and efficiency. As annotated 4D datasets expand and more sophisticated modeling strategies emerge, we expect temporal indoor segmentation methods to become increasingly robust and scalable.

\temporalJointFigure{scene0079}{00}{08}{fig:merging_stools}{%
  \textbf{Temporal Segmentation Comparison---Baseline Comparison.} Mask4former merges the two stools as one instance in a patched manner and considers the sofa as two different instances, despite predicting the same class label. Mask3D produces several semantic segmentation errors between the two scans (and the ground truth), hence leading to incorrect temporal association of instances. \project{} robustly associates semantics and instances across time even if classifying a stool as a sofa.
}

\temporalJointFigure{scene0449}{00}{02}{fig:partial_overlap}{%
  \textbf{Temporal Segmentation Comparison---Baseline Comparison.} Errors in the Mask3D baseline arising from partial observations can be corrected by incorporating temporal information; \project{} reliably identifies the partially visible coffee table, as shown in both instance and semantic segmentation predictions. Mask4Former also benefits from temporal overlay when there are no object changes, relying heavily on spatial alignment. All methods struggle with inconsistently annotated couches and pillows in 3RScan, where labels are often interchangeable between these objects across different scenes in the dataset. 
}

\temporalInstanceFigure{scene0189}{02}{03}{fig:ball}{%
  \textbf{4D Semantic Instance Segmentation---Baseline Comparison.} Mask4Former fails to identify moved chair identities when there is no spatial overlap and confuses the ball mask due to overlapping spatial regions between stages. Mask3D fails to correctly associate one of the chairs. In contrast, \project{} accurately tracks and masks the moving chairs.
}

\temporalInstanceFigure{scene0459}{00}{01}{fig:ambiguities_challenge}{%
  \textbf{4D Semantic Instance Segmentation---Baseline Comparison.} Ambiguous scenarios, with a set of objects with similar features (here \textit{chairs}), present a challenge and remain a limitation for all 4DSIS methods. In this example, Mask4Former merges the identities of two chairs into one instance in the first scan while still tracking a single chair in the next scan. Mask3D looses track of two chairs in the temporal matching and fails to match static instances such as the table and the curtains. \project{} primarily maintains correct identities, with only a single missed chair tracking.
}

\temporalInstanceFigureOursBeside{scene0079above}{06}{10}{fig:no_overlap}{%
  \textbf{4D Semantic Instance Segmentation---Without Spatial Overlap.} In scans of the same scene that have zero overlap, a common ceiling instance is correctly identified by \project{} without forcing matches between non-spatially overlapping objects.
}

\temporalInstanceFigureSharing{scene0170sharing}{00}{01}{fig:sharing}{%
  \textbf{4D Semantic Instance Segmentation Architecture Ablation.} Comparison between our best-performing \project{}, which shares information temporally via cross-time contrastive loss and ST-masking (left), and a standard spatio-temporal architecture that decodes joint queries without explicit temporal information sharing. While the standard method benefits from unified 4D modeling, it struggles with ambiguous instances—e.g., incorrectly segmenting two chair seats as one due to reliance on semantic consistency. Explicit temporal information sharing helps disambiguate instances by introducing geometric and contrastive cues without rigid constraints.
}

\temporalInstanceFigureOurs{scene0180}{00}{01}{fig:bathroom}{%
  \textbf{4D Semantic Instance Segmentation---Dataset Annotation.} Here we illustrate ground truth annotation inconsistencies, particularly for small instances, where \project{} predicts objects not labeled and missing an annotated object. For example, the soap bottle on the sink is annotated in the ground truth but missed by us, vs. the shampoo bottle on the right corner of the bathtub is correctly and consistently segmented by us but is non-existent in the ground truth. We are thus penalized in both cases.
}

\temporalInstanceFigure{scene0009}{01}{03}{fig:ambiguities_metric}{%
  \textbf{4D Semantic Instance Segmentation---Metric Highlight.} All methods correctly associate ambiguous chairs, but Mask3D visually shuffles identities within the group due to the lack of geometric information, not matching chairs to their nearest neighbor even though they might have remained static. Our metric, t-mAP, does not penalize such cases, as this represents a valid solution.
}

\end{document}